\pdfoutput=1

\documentclass[10pt,twocolumn]{article}
\usepackage{simpleConference}
\usepackage[lined,boxed,commentsnumbered, ruled]{algorithm2e}
\usepackage{times}
\usepackage{graphicx,subfigure, amsmath}
\usepackage{amssymb}
\usepackage{url}
\usepackage{multirow}
\usepackage{tikz}
\usepackage{amsmath}
\usepackage{url}
\usepackage{caption}
\usepackage{gensymb}

\begin{document}

\title{Self-Exploration in Complex Unknown Environments using Hybrid Map Representation}

\author{Wenchao Gao, Matthew Booker, Jiadong Wang
\thanks{W. Gao is with Institute for Infocomm Research, Singapore. Email: gaowc1990@gmail.com. M. Booker is with University of California, Irvine, USA. J. Wang is with National University of Singapore, Singapore.}
}

\maketitle
\thispagestyle{empty}

\begin{abstract}
A hybrid map representation, which consists of a modified generalized Voronoi Diagram (GVD)-based topological map and a grid-based metric map, is proposed to facilitate a new frontier-driven exploration strategy. Exploration frontiers are the regions on the boundary between open space and unexplored space. A mobile robot is able to construct its map by adding new space and moving to unvisited frontiers until the entire environment has been explored. The existing exploration methods suffer from low exploration efficiency in complex environments due to the lack of a systematical way to determine and assign optimal exploration command. Leveraging on the abstracted information from the GVD map (global) and the detected frontier in the local sliding window, a global-local exploration strategy is proposed to handle the exploration task in a hierarchical manner. The new exploration algorithm is able to create a modified tree structure to represent the environment while consolidating global frontier information during the self-exploration. The proposed method is verified in simulated environments, and then tested in real-world office environments as well.
\end{abstract}

\section{Introduction}
\label{introduction}

Traditional robotic navigation requires a known or predefined map before navigation goals can be determined and executed by the motion planner \cite{Bruemmer2009}. As the robotics industry grows rapidly, the ability to investigate and operate independently in an unknown environment becomes essential for an advanced robot to be considered fully autonomous. According to \cite{frontier1997}, self-exploration and mapping can be defined as the action of autonomously moving through an unknown environment while building a map that can be used for subsequent navigation.

In literature, solutions for self-exploration in unknown environments have been reported and divided into two categories: randomized-based searches \cite{random_walk, greedy_mapping, Oriolo2004,Umari2017} and frontier-driven strategies \cite{frontier1997, Banos2002, Keidar2012, Senarathne2013}.

In the first category, straightforward approaches employ randomized selection mechanisms \cite{random_walk} or greedy based searches \cite{greedy_mapping} to explore the environment. Although simple and fast, such strategies yield locally optimal solutions but do not guarantee global optimization in many cases. To address the issue, Sensor-based Random Tree (SRT) method \cite{Oriolo2004}, which can be considered as a goal-oriented exploration strategy, bias the randomized generation of configurations towards unexplored areas. However, these approaches suffer from the problem of revisiting explored places. Recently, a new exploration strategy leveraging on Rapidly-exploring Random Trees (RRT) utilizes the randomized tree expansion to detect and prioritize unknown spaces \cite{Umari2017}. RRT techniques ensure complete search coverage and can be extended to higher dimensions, but result in a lower exploration efficiency when searching in complex spaces, such as office areas with narrow corridors \cite{tencon2017}.

More efficient approaches make use of the concept of map frontier. The key idea of this branch is to determine the next desired goal based on frontiers, i.e. boundaries between the known and unknown cells in an occupancy grid map. In the pioneer work of \cite{frontier1997}, frontier edges are required to be segmented from a dynamical occupancy grid map in order to determine potential targets. The selected target will be assigned as a temporary destination point. To improve the frontier detection efficiency, in \cite{Banos2002} and \cite{Keidar2012}, a series of target points in the grid map reveal the quality of the candidate points around a frontier which will be evaluated according to some criteria. Senarathne et al. develop an efficient approach to segment frontiers by only detecting intermediate changes to cells in the current exploration map and only the updated grid cells are considered for the frontier segmentation \cite{Senarathne2013}.

To produce accurate maps, metric or grid-based SLAM techniques are frequently incorporated with frontier-driven exploration \cite{metric1,metric2}. However, these approaches usually have to process the entire map to detect the desired frontier. If the map is dynamically updated and becomes larger, more computational resources and memories are required \cite{Keidar2012}, which often prohibits the exploration efficiency in large-scale environments. Another issue with the current frontier-driven methods is that they usually have poor capability to efficiently select and assign frontier in a systematic way when the search area is large and cluttered, resulting in back and forth exploration over visited places in a complex searching space.

Some other approaches that leverage the use of a topological map \cite{topological1,topological2} have been proposed to represent the unknown environment in a qualitative manner. A local and global decision-making mechanism for self-exploration is proposed in \cite{topological1}, where a bubble searching mechanism based on local geometric features is used to determine robot orientation and a topological map is built to move the robot between different topological nodes globally. The problem of high computational cost in large-scale environments can be alleviated, and yet a place recognition algorithm is a prerequisite for this method, making it vulnerable in places that are difficult to be recognized. The topology of the environment is encoded in a Generalized Voronoi Diagram (GVD) in \cite{topological2}. The GVD containing the key geometric information can be interpreted as an efficient topological representation of an indoor/outdoor environment. However, in return, the topological methods lose the metric property and may encounter the problem of ambiguous spatial reasoning between topological classes.

To further improve the exploration efficiency, recent works \cite{hybrid1,hybrid2} propose a hybrid map representation using metric and topological information. In \cite{hybrid1}, an ear-based exploration strategy makes use of GVD-based topological graph and extended Kalman filter (EKF) to track the pose of the robot. The ear-based strategy is reported to facilitate loop closure in SLAM process, assuming that several small obstacles exist in the search space. An incrementally constructed GVD for frontier-based exploration is introduced to completely solve the pose-SLAM problem in \cite{hybrid2}. However, the proposed diagram carries redundancies resulting in chaotic exploration decisions which degrade exploration efficiency.

\section{Motivation and Overall Strategy}
\label{motivation}

The goal of work is to develop an efficient self-exploration navigator which maximizes the mapping coverage as quickly as possible in an unknown environment. To gather the local metric information efficiently, a modified frontier-based method is proposed to realize a stem-first exploration. The frontier method is employed due to its efficiency in local unexplored space searching and ease of integration with the grid-based SLAM techniques such as \cite{Slam2005}.

Considering the limitation of the current frontier methods, the concept of topology is introduced to consolidate all frontier information from a global vision and systematically determine optimal unexplored places for the mapping agent. Different from the existing methods, the working space is categorized into two parts: ``Stem" and ``Branches". The region of ``Stem" can be considered as the main road in a metric map or the backbone in a topological way, while ``Branches" are rest scattered areas. By taking the robot heading information and map topology into account, the proposed strategy prefers to navigate along the ``stem" to explore the main structure of the space first. And then prioritize the unexplored spaces (Branches) based on a global decision making. The global call will be activated to choose an optimal area to explore when the ``stem" has been fully explored or the robot change its orientation rapidly in a cross-road or dead-end.

Borrowing the idea of hybrid map representation \cite{hybrid1,hybrid2}, an innovative hierarchical exploration algorithm is proposed in this paper. The hierarchical strategy has been designed in a global-local-cooperative fashion. More specifically, in a lower level control, the local desired frontiers pushing the robot to stay on the main road are determined and assigned to the navigator within a sliding local window. Globally, a GVD-based topological planner taking the role of an upper level decision maker is developed to abstract the metric information of all global frontiers through a modified tree structure named as multi-root tree. It is noteworthy that the hierarchical strategy is proposed to achieve a systematic way of exploring complex unknown environments by combining the benefits of both metric and topological map information.

\section{Preliminary Terminology}
\label{terminology}
In this section, we provide the definition of functions and symbols related to the proposed approach.

\textbf{Occupancy Grid}: The representation of a map that divides the space into grid cells.

\textbf{Search Space} $\mathbb{R}^2$: The set of the whole search space. This set in $2D$ consists of free $\mathbb{R}_f$, occupied $\mathbb{R}_o$, and unknown space $\mathbb{R}_u$, i.e, $\mathbb{R}=\mathbb{R}_f\cup\mathbb{R}_o\cup\mathbb{R}_u$

\textbf{Frontiers} $\mathcal {F}$: A list $\mathcal {F}=\{f_0,...,f_j \}$ that stores all nearby frontier nodes. The desired frontier $f^* \in \mathcal {F}$ will be assigned as the exploration goal.

\textbf{Utility Cost} $C$: This cost function is defined to determine the most desirable frontier $f^*$ to be explored from list $\mathcal {F}$.

\textbf{Topological Node} $N$: A set of nodes $N=\{\nu_0,...,\nu_k\}$ denoting the location of a GVD vertex. Nodes along the main path are called ``stem nodes", the others located in the branches of the GVD graph are named ``branch nodes".

\textbf{Edge} $E$: An edge linking two topological nodes. Edges are divided into two categories: edges between two stem nodes $\eta\in E$ and edges connecting to the branch nodes $\epsilon\in E$.

\textbf{Topological Map} $G$: A graph-based map constructed by edges and topological nodes, i.e. $G=(N,E)$.

\begin{figure}[tb]
	\centering{
	\includegraphics[width=2in]{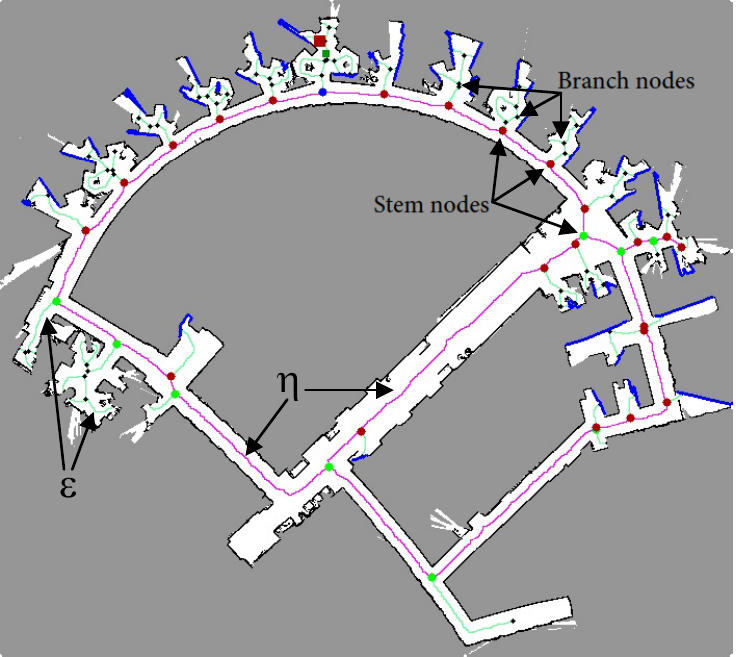}}
	\caption{An example of GVD map with detected frontiers.}
	\label{graphic_example}
    \vspace{-0.4cm}
\end{figure}

A graphic example showing the hybrid map representation is illustrated in Fig. \ref{graphic_example}, containing the detected frontiers and the topological map in an office area. The frontiers in $\mathcal {F}$ are highlighted by blue boundary lines for all unexplored areas. Stem nodes are denoted by red (linked to frontiers) and green (not directly linked to frontiers) dots. Branch nodes are highlighted in black. The map $G$ is connected by red edges ($\eta$) along the main path and green edges ($\epsilon$) at the branches.

\section{Methodology}
\label{methodology}

Assuming that the environment can be represented in a topological form, the main idea of the proposed strategy is to take a backbone traversal by examining the topological stem and then explore the remaining areas at the branches. A cycle of two stages of decision making is designed to implement the idea, i.e. local frontier detector and GVD topological planner. The two stages cooperate with each other in a hierarchical way. The lower level stage is to obtain new information by moving the robot to the boundaries between open space ($\mathbb{R}_f$) and unknown space ($\mathbb{R}_u$) inside a sliding window. Metric information measured by laser is used to build up environmental structures and detect local frontiers. The upper level stage, leveraging on topology, deals with the global exploration planning when local information is not reliable.

\subsection {Hierarchical Exploration Strategy}

\begin{figure}[tb]
	\centering{
	\includegraphics[width=3in]{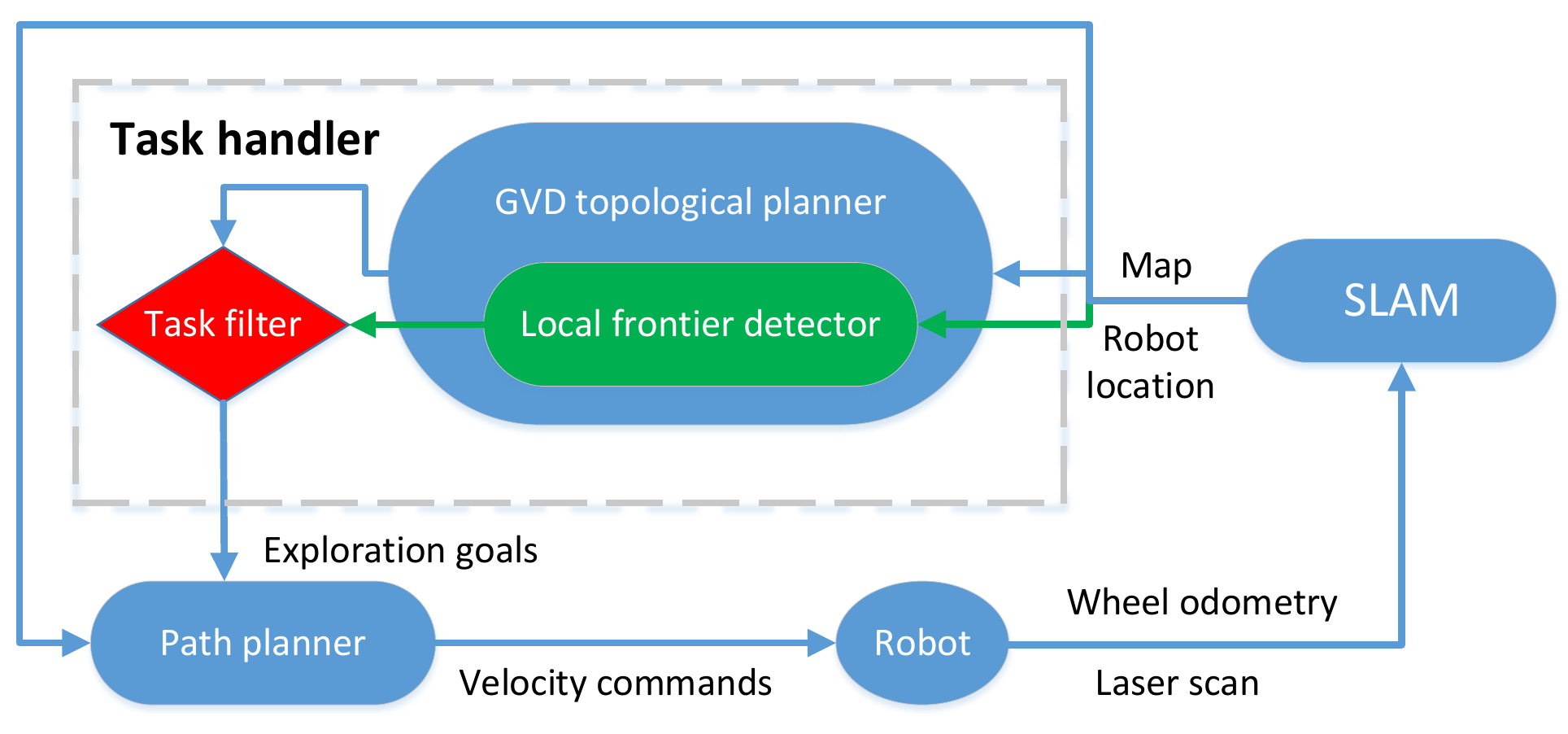}}
	\caption{Overall schematic diagram of the hierarchical exploration strategy.}
    \vspace{-0.4cm}
	\label{framework}
\end{figure}

Fig. \ref{framework} shows the framework of the hierarchical exploration strategy. As can be seen, a pose SLAM method  named ``Karto" \cite{karto} is deployed to map multiple unexplored areas. Taking odom data and sensor information as inputs, it produces a  metric map and the robots location for the task handler to generate exploration goals. Each goal generated by the task handler will be assigned to the path planner, resulting in a series of velocity commands to drive the robot into new territory.

\begin{algorithm}
{
\caption{Task handler for exploration goal generation}
\begin{enumerate}
\item[1.] \textbf{while} $\emph{HierarchicalPlanner}=1$ \textbf{do}
\item[2.] \qquad run \emph{LocalFrontierDetector}
\item[3.] \qquad \textbf{if} ($\mathcal {F}=\emptyset$) \textbf{or} ($\emph{angleChanged}>=150 \degree $) \\ \qquad \textbf{then} \qquad \qquad \qquad \qquad \qquad \qquad $\rhd task \ filter$
\item[4.] \qquad \quad run \emph{GVDTopologicalPlanner}
\item[5.] \qquad \quad \textbf{if} ($v^*\neq \emptyset$) \textbf{then}
\item[6.] \qquad \qquad $goal\leftarrow v^*$
\item[7.] \qquad \qquad run \emph{SendGoal(goal)}
\item[8.] \qquad \qquad \textbf{continue}
\item[9.] \qquad \quad \textbf{else if}   ($v^* =\emptyset$) \textbf{and} ($\mathcal {F}=\emptyset$)
\item[10.] \qquad \qquad  $\emph{HierarchicalPlanner}=0$  \qquad \quad $\rhd end \ task$
\item[11.] \qquad \textbf{else}
\item[12.] \qquad \quad $goal\leftarrow f^*$
\item[13.] \qquad \quad run \emph{SendGoal(goal)}
\end{enumerate}
\label{hierarchial}
}
\end{algorithm}

Our task handler is summarized in \textbf{Algorithm \ref{hierarchial}}. Two planning stages (`\emph{LocalFrontierDetector'} and `\emph{GVDTopologicalPlanner}') are designed in a hierarchical way that the robot prioritizes local frontier searching in a sliding window (\emph{line 2}), and makes upper-level decision by requesting the GVD topological map (\emph{line 4}) based on the conditions specified by the task filter (\emph{line 3}). The task filter will trigger `\emph{GVDTopologicalPlanner}' and deactivate `\emph{LocalFrontierDetector}' when either of the two conditions are satisfied: 1) no local frontier is detected; 2) the robot changes its orientation significantly in a short period of time (2 seconds). `\emph{LocalFrontierDetector}' will be activated again once the GVD exploration goal is reached.

Only one exploration goal from either stage will be activated for each iteration to prevent sending multiple commands to the robot. The filtered exploration goals are assigned to the path planner to obtain velocity commands to actuate the robot (\emph{line 7 and 12}). The exploration task terminates when both planning stages return \emph{NULL} (\emph{line 9}).
It is proven by simulation and experimental results that the hierarchical task handler is able to effectively combine the two planning stages by taking advantage of both metric and topological information. The way to determine the desired frontier $f^*$ and the GVD exploration goal $v^*$ will be discussed in the following two subsections.

\subsection {Oriented Local Frontier-driven Exploration}

Within the occupancy grid any unknown cells adjacent to free cells are grouped together into regions. The centroid of each region (above a certain minimum size) can be considered as a frontier node $f_j$. The frontier list $\mathcal {F}_t$ contains all the valid frontier nodes at time $t$.

The most widely used frontier-driven approaches \cite{frontier1997,Umari2017,exploration2007} determine the desired frontier $f^*$ by taking into consideration the frontier size and distance, which also has been referred as the greedy frontier-driven exploration. More specifically, in these greedy approaches $f^*$ is selected by minimizing the following utility function:
\begin{equation}
\label{nsingsta}
C(f_j) = (\omega_d \times {f^D_j} - \omega_s \times {f^S_j})  \nonumber
\end{equation}
\begin{equation}
\label{old_cost}
f^* = \emph{Arg}\operatorname*{min}_{f_j \in \mathcal {F}} \Bigg(C(f_j)\bigg)
\end{equation}
where $f^D_j$ is the Euclidean distance from the robot to the frontier node and $f^S_j$ is the grid size of the frontier area. $\omega_d, \omega_s$ are weighting parameters associated with the two terms. By minimizing the utility function, the robot takes the shortest path from its current location to the boundary containing the most unknown information. To be noted that the optimal frontier $f^*$ is selected among all the detected frontiers in the global map every iteration.

It has been reported in \cite{frontier1997} that by constantly moving to new frontiers, the robot is able to extend its map into new space until the entire environment has been explored. However, Eq. 1 can be inefficient when the unknown environment is complex and dynamic (e.g, office area with narrow corridors, secluded cubicles, and possible moving pedestrians) as it only takes into account Euclidean distance as opposed to actual travel distance. In cases where a frontier is behind a large object, such as a wall, the Euclidean distance is lower than the travel distance. The cluttered environment thus causes the robot to travel back-and-forth repeatedly over explored locations reducing the exploration efficiency. To avoid this, an efficient way of measuring the travel distance from the robot position to the candidate frontiers must be developed. Standard path planning techniques such as A* searching and RRT can solve the problem yet at a higher computational cost such as \cite{Umari2017}. The following frontier detection mechanism combining with the topological tree representation has the ability solve the back-and-forth trap efficiently.

A new utility cost $\widetilde{C}$ is designed to incorporate the frontier orientation information in `\emph{LocalFrontierDetector}'. As a result, the modified frontier detector ensures exploring in a certain direction as far as possible before turning or backtracking. The utility cost $\widetilde{C}(\widehat{f}_j)$ can be written as:
\begin{equation}
\label{new_cost}
\widetilde{C}(\widehat{f}_j) = (\|\widehat{f}^D_j\| - \|\widehat{f}^S_j\| + \|\widehat{f}^R_j\| )
\end{equation}
where, $\widehat{f}^D_j, \widehat{f}^S_j$ are defined the same as Eq. (\ref{old_cost}). $\widehat{f}^R_j$ denotes the steering angle to face each frontier node. And the candidate for desired frontier has been narrowed down to those frontiers $\widehat{f}_j$  within a certain distance around the robot (local window) to speed up the searching process (global information will be handled by the global decision maker). The three cost components are normalized into the range of $[0,1]$ to balance the overall utility cost and omit the process of parameter selection. The scaled values $ \|\widehat{f}^D_j\|$, $\|\widehat{f}^S_j\|$, $\|\widehat{f}^R_j\|$ are computed as below:
\begin{equation}
\label{normalization}
\|\widehat{f}^D_j\|=\frac{\widehat{f}^{D}_j - \widehat{f}^{D}_{min}}{\widehat{f}^{D}_{max} - \widehat{f}^{D}_{min}}
\nonumber
\end{equation}
\begin{equation}
\label{normalization}
\|\widehat{f}^S_j\|=\frac{\widehat{f}^S_j - \widehat{f}^S_{min}}{\widehat{f}^S_{max} - \widehat{f}^S_{min}},\ \|\widehat{f}^R_j\|=\frac{\widehat{f}^R_j - \widehat{f}^R_{min}}{\widehat{f}^R_{max} - \widehat{f}^R_{min}}
\end{equation}

The ``min" and ``max" sign indicate the minimal and maximum value of each cost component. The local desirable frontier node $\widehat{f}^*$, thus, can be determined when $\widetilde{C}(\widehat{f}_j)$ is minimized as:
\begin{equation}
\label{nsingsta}
\widehat{f}^* = \emph{Arg}\operatorname*{min}_{\widehat{f}_j \in \mathcal {F}} \Bigg(\widetilde{C}(\widehat{f}_j)\bigg)
\end{equation}
		
Considering a 2D navigation scenario, the pose associated with the desired frontier $\widehat{f}^*(x,y,\theta)$ is the output of exploration stage `\emph{LocalFrontierDetector}'.

\subsection {GVD-based Topological Planner}

The GVD graph representation is used to obtain the topological structure of the grid map \cite{gvd}. An common approach is to create a GVD-Matrix $M$ that can indicate whether a cell belongs to GVD. To get the topological map $G$, image processing techniques are applied to recognize the intersection points in $M$. These points are stored as nodes in the topological node set $N$. Moreover, points of GVD which cross frontiers will be attributed to $N$ as well. Once all the topological nodes are detected, the edges $E$ are generated by gathering the GVD's cells between nodes using a point queue.

One novelty of this paper is that a modified version of the tree data structure called multi-root tree is introduced to represent the graph. Within set $N$, each stem node can be regarded as a root of a binary tree, the other nodes are treated as branch nodes. Branch nodes connecting to frontiers are defined as leaf nodes. They can be traced back to the corresponding root in the same tree. Therefore, the graph can be considered as a combination of multiple tree structures. By consolidating all the roots, the topological planner is able to access the abstracted frontier information of the whole graph and make exploration decisions at the global level.

Based on the concept of a multi-root tree, the planner can be divided into four steps: 1) Determining the stem and branch nodes of the GVD graph; 2) Transforming the GVD graph into a multi-root tree and tracing all frontier information to the roots; 3) Finding the nearest root $v^{key}$ to the current robot pose; 4) Determining the best stem node as the exploration goal $v^*$, based on a specific score function.

\begin{algorithm}
{
\caption{DFS-based main path searching}
\begin{enumerate}
\item[] \textbf{$MainPathSearch(v^c)$}
\item[] \textbf{Input}: a node $v^c$.
\item[] \textbf{Output}: $L^c$ the longest path length from $v^c$, $N^c$ the corresponding set of nodes.
\item[] \textbf{Initialization}: boolean type vector $Visited$ that indicates whether a node has been visited.
\item[1.] $Visited[v^c]$ $\xleftarrow{}$ 1.
\item[2.] $L^c$ $\xleftarrow{}$ 0.
\item[3.] $N^c$ $\xleftarrow{}$ empty vector.
\item[4.] \textbf{if} $Neighbour^c = \emptyset$ \textbf{then}
\item[5.] \qquad \textbf{return} $(L^c, N^c)$
\item[6.] \textbf{for} $v^n: Neighbour^c$
\item[7.] \qquad \textbf{if} $Visited[v^c] \neq 1$
\item[8.] \qquad \qquad $(L^n, N^n)$ $\xleftarrow{}$ $MainPath Search(v^n)$
\item[9.] \qquad \qquad $TempL^c$ $\xleftarrow{}$ $L^n + L^{nc}$
\item[10.] \qquad \qquad \textbf{if} $TempL^c > L^c$
\item[11.] \qquad \qquad \qquad \textbf{then} $(L^c, N^c)$ $\xleftarrow{}$ $(TempL^c, N^n)$
\item[12.] \qquad \qquad \textbf{else if} $TempL^c > threshold^l $
\item[13.] \qquad \qquad \qquad \textbf{then} \textbf{add} $N^n$ into $subN$
\item[14.] \textbf{add} $v^c$ into $N^c$
\item[15.] \textbf{return} $(L^c, N^c)$
\end{enumerate}
\label{gvd_dfs}
}
\end{algorithm}

Seperating GVD nodes into stem and branch nodes is a vital step. A depth-first-search (DFS) algorithm is developed in a recursive way as shown in \textbf{Algorithm \ref{gvd_dfs}}. $Neighbour^c$ is the set of nodes that directly connect to the current node. $v^n\in Neighbour^c$ is one of the neighbouring nodes to the current node $v^c$. $L^{nc}$ is the length of the edge connecting $v^c$ and $v^n$. The recursive mechanism (\emph{line 6 to line 13}) enables the algorithm to search and store the longest path of nodes ($N^c$) which is considered the main path ($L^c$) in the graph. The results can be seen in Fig.\ref{graphic_example}, where the stem and branch nodes are properly clustered into two groups.

\begin{figure}[htb]
	\centering{
	\includegraphics[width=2.5in]{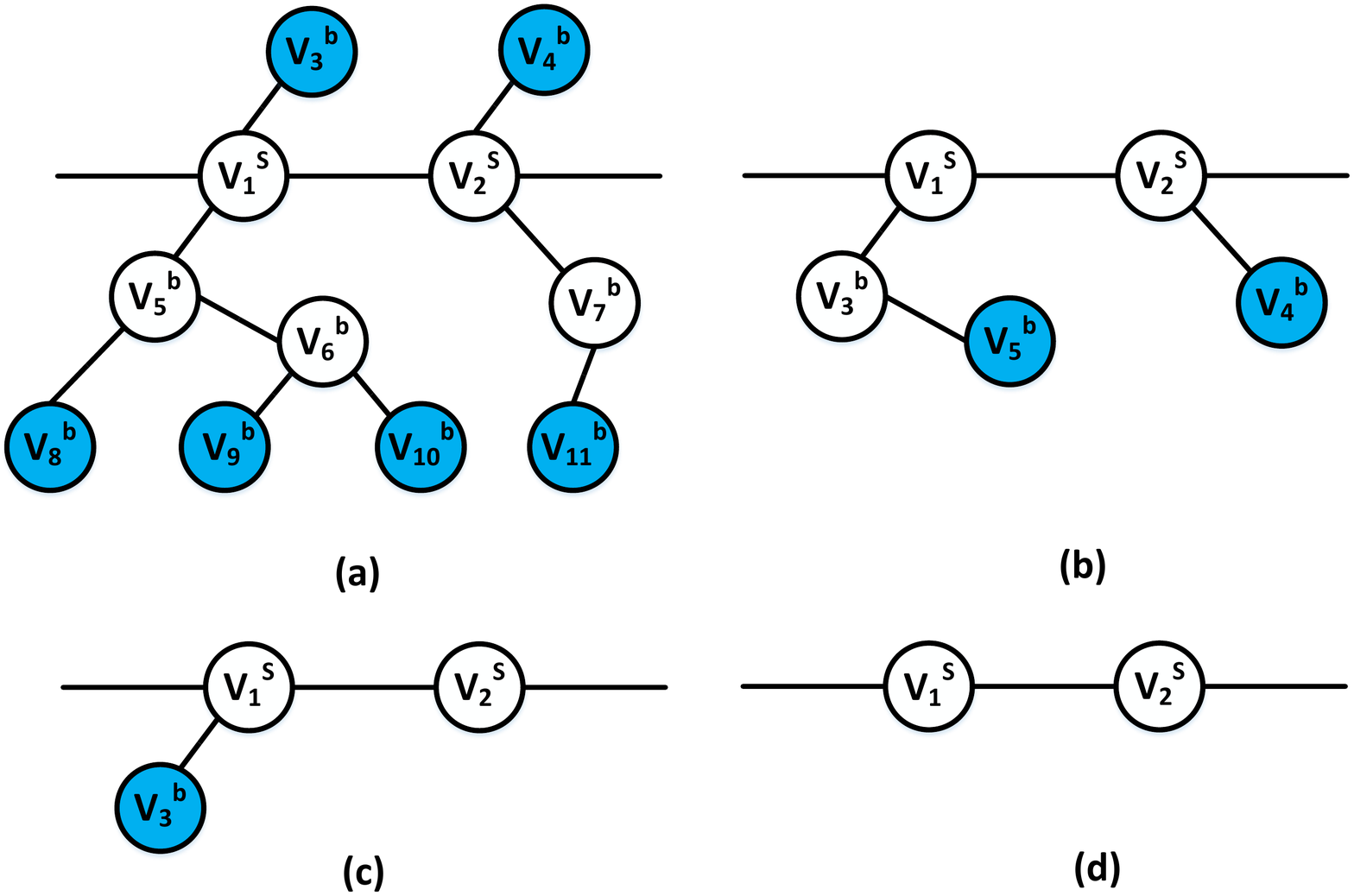}}
	\caption{The construction of a multi-root tree: (a) Original graph, (b)-(c) Fusion process and information abstraction, (d) Multi-root tree construction finished.}
    \vspace{-0.2cm}
	\label{refine}
\end{figure}

In order to abstract the frontier information from leaf nodes, Fig.\ref{refine} illustrates how the multi-root tree is constructed. The original topological graph is shown in (a), where $V^s_1$ and $V^s_2$ are stem nodes and also the roots of two tree structures. All the leaf nodes labelled as $V^b$ need to be backtracked to $V^s$. From (b) to (c), the blue nodes are fused into their parents nodes and transmit the frontier information up to the root level layer by layer. When no more blue nodes can be fused, the process is finished as shown in (d). By performing step 2), all information at the branches is transmitted to the root so that the processed stem nodes are able to represent the whole unexplored space. In the proposed multi-root tree structure all stem nodes are at the same root level with no parent node. This structure tackles the ordering problem seen in other tree structures that occurs when the stem nodes create a cycle.

\begin{figure}[htb]
	\centering{
	\includegraphics[width=2.5in]{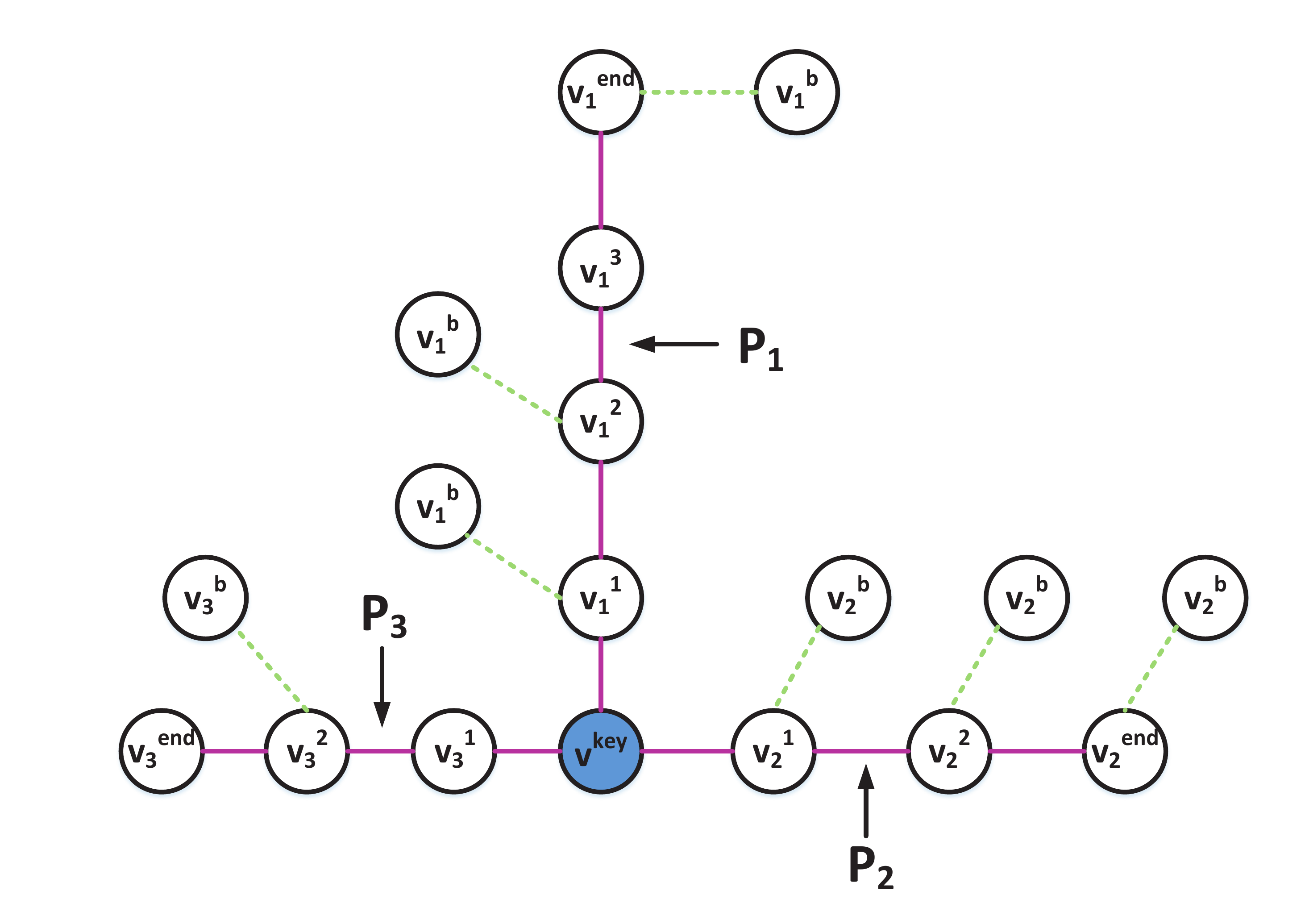}}
	\caption{Diagram of choosing optimal path and starting node.}
    \vspace{-0.4cm}
	\label{scorediagram}
\end{figure}

Next, in order to make a global decision, the robot pose is taken as a reference to search for the nearest stem node $v^{key}$, highlighted in blue in Fig. \ref{scorediagram}. For instance, there are three possible paths, $P_1$, $P_2$ and $P_3$. The starting nodes of these three paths are $v_1^{1}$, $v_2^{1}$ and $v_3^{1}$, respectively. Similarly the most distant stem node of each path is denoted as $v_1^{end}$, $v_2^{end}$ and $v_3^{end}$. All branch nodes $v^b$ are fused into the root nodes through the previous steps. For the sake of efficiency in global planning, all stem nodes along each path are taken into consideration when determining the exploration goal. Thanks to steps 1) and 2), the processed stem nodes containing frontier information can be used to design a score function that evaluates individual node scores. By summing the node scores in one path we can obtain the total score for that path. The $NodeScore$ and $PathScore$ functions are expressed as:
\begin{equation}
\label{scorefunction}
Node Score(v^j) = \sum_{i=1}^k S_{i} \cdot e^{-d^j}
\end{equation}
\begin{equation}
\label{scorefunction}
Path Score = \frac{\sum_{j=1}^n Node Score(v^j)}{log(l+1)}
\end{equation}
where, $k$ is the number of frontiers associated with $v^j$. $S_i$ is the size of the $ith$ frontier associated with $v^j$. $d^j$ is the real world travel distance between $v^{key}$ and $v^j$, obtained by GVD. $n$ represents the number of nodes within the current path. While $l$ can be considered as the approximated real world travel distance to the last stem node in the path ($v^{end}$). As described before, all GVD nodes $v\in N$ are equidistant to nearby obstacles \cite{gvd}. By taking advantage of this property, the travel path between two nodes has a great ability to avoid obstacles and its distance is shortest. Hence, we can approximate travel distances by counting the number of GVD points between two nodes and multiplying it by the resolution of the corresponding grid map (measuring the length of Edge $E$ between topological nodes as shown in Fig. \ref{graphic_example}), instead of applying an extra path planning algorithm.

By incorporating $l$ and $d^j$, the global planner avoids the back-and-forth trap by selecting stem nodes based on their approximate travel distances. The optimal path $P^*$ is the one with the highest $PathScore$ value, which takes all the global frontier information into account. Simultaneously, the first node of path $P^*$ is chosen as the exploration goal $v^*$, which is also the output of the global planning stage. To be noted, the topological planner, though takes extra computation, is activated only in certain conditions (\textbf{Algorithm \ref{hierarchial}}). The testing results show that the overall processing speed will not be significantly affected.

\section{Experiments}
\label{experiment}

\subsection {Simulation Results}
\begin{figure}[tb]
\centering{
\subfigure[$Sketched_1$]{\includegraphics[width=1.55in,height=1in]{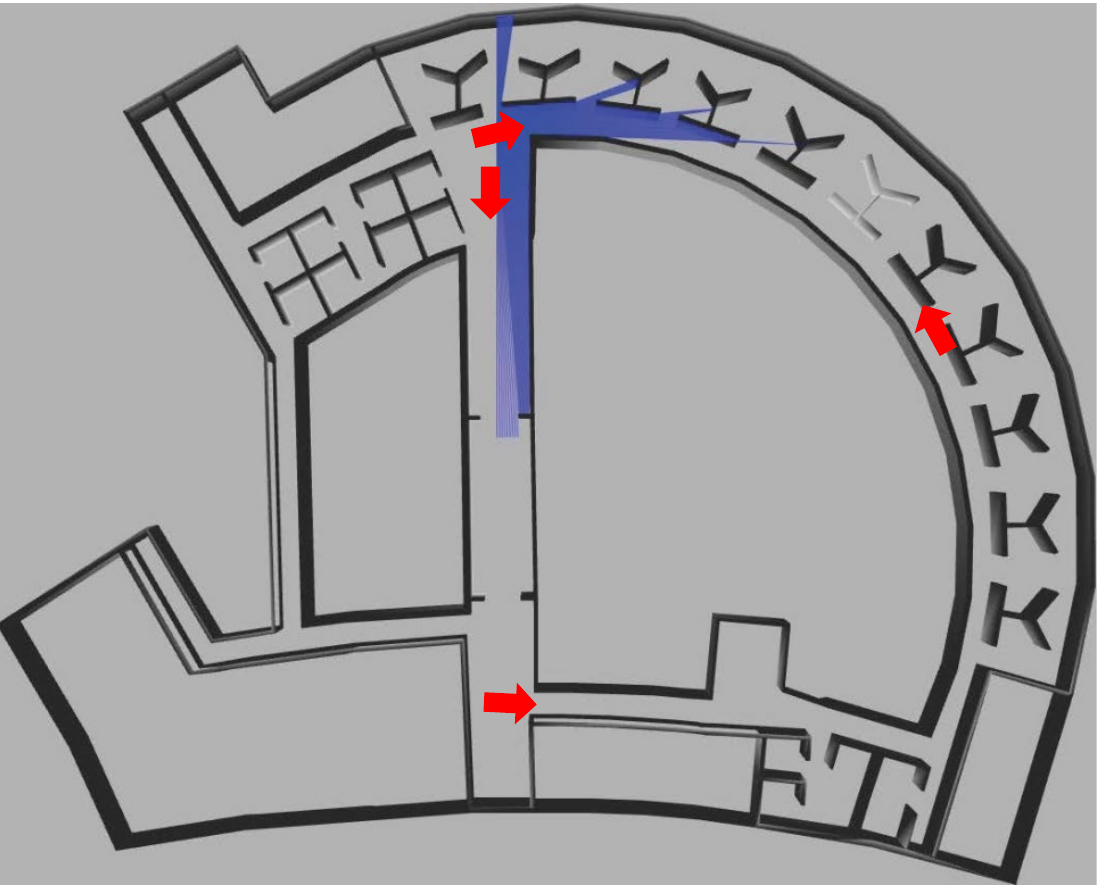}}
\subfigure[$Sketched_2$]{\includegraphics[width=1.55in,height=1in]{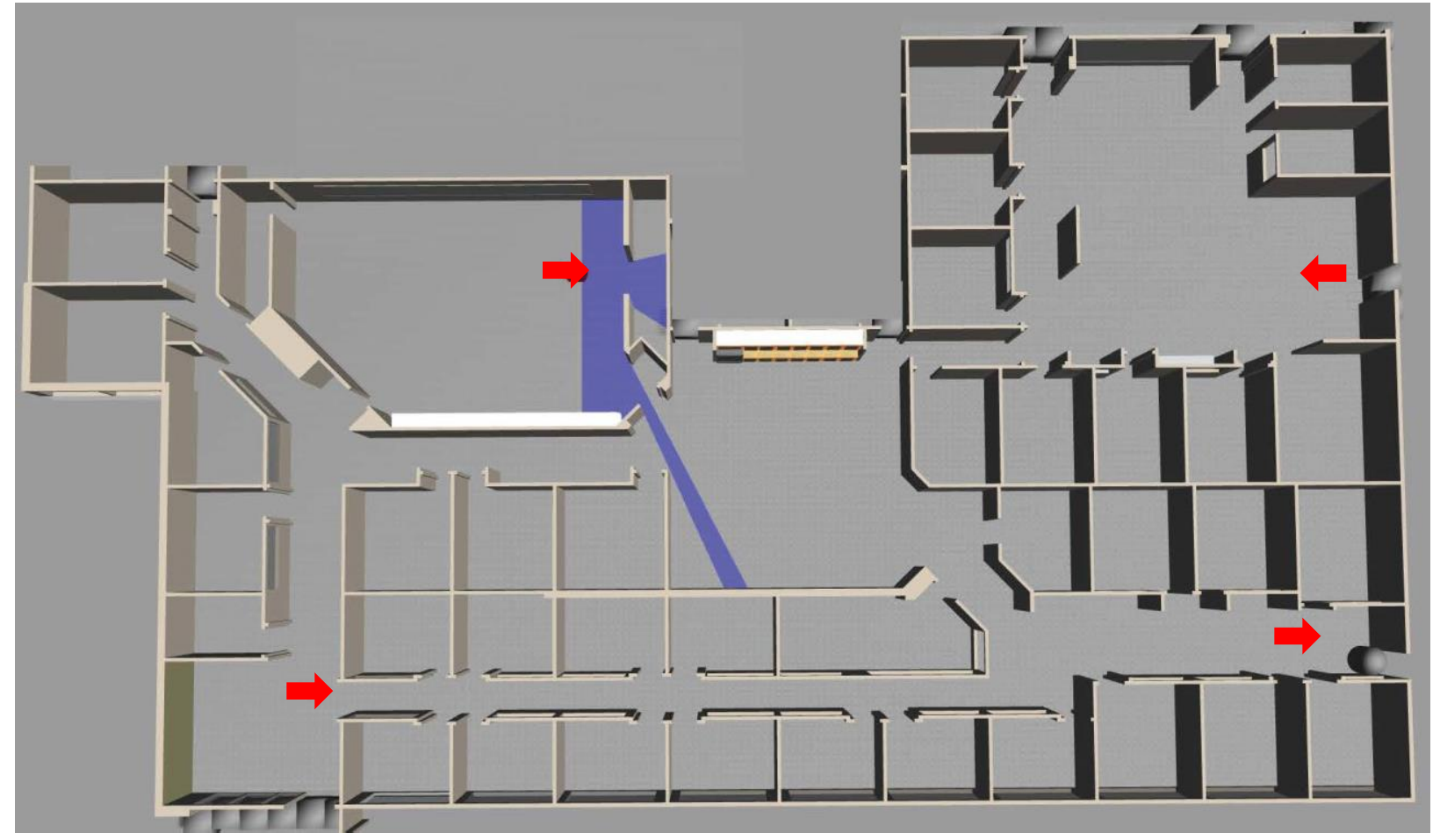}}
}
\caption{Simulation environments for exploration test. Red arrows indicate different initial poses for the robustness test.}
\vspace{-0.2cm}
\label{gazebo}
\end{figure}

To evaluate the proposed algorithm, two sketched map models shown in Fig. \ref{gazebo} are generated in Robot-Operating-System (ROS) Gazebo simulator. The size of two maps are listed in Table \ref{comparison}. The simulator can generate realistic robot movement and sensor data combined with noise. The proposed exploration algorithm is implemented in ROS environment, and compared against the open-source greedy frontier-driven exploration method \cite{online} and the RRT detector in \cite{Umari2017}. Off-the-Shelf ROS packages are used in the implementation of SLAM (Karto) and motion planning (ROS Navigation Stack).

\begin{figure*}[htb]
\centering{
\subfigure[Time stamp: 12.5s ]{\includegraphics[width=1.65in,height=1.3in]{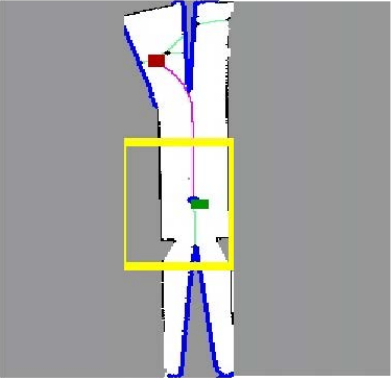}}
\subfigure[Time stamp: 111.3s]{\includegraphics[width=1.65in,height=1.3in]{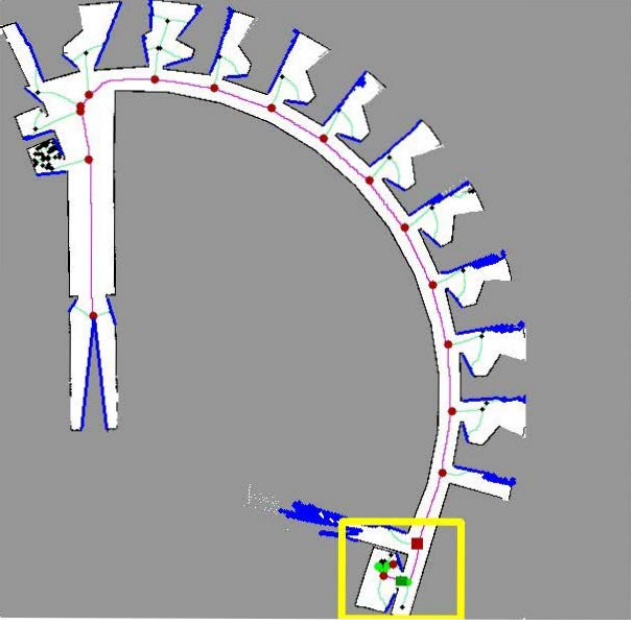}}
\subfigure[Time stamp: 312.8s]{\includegraphics[width=1.65in,height=1.3in]{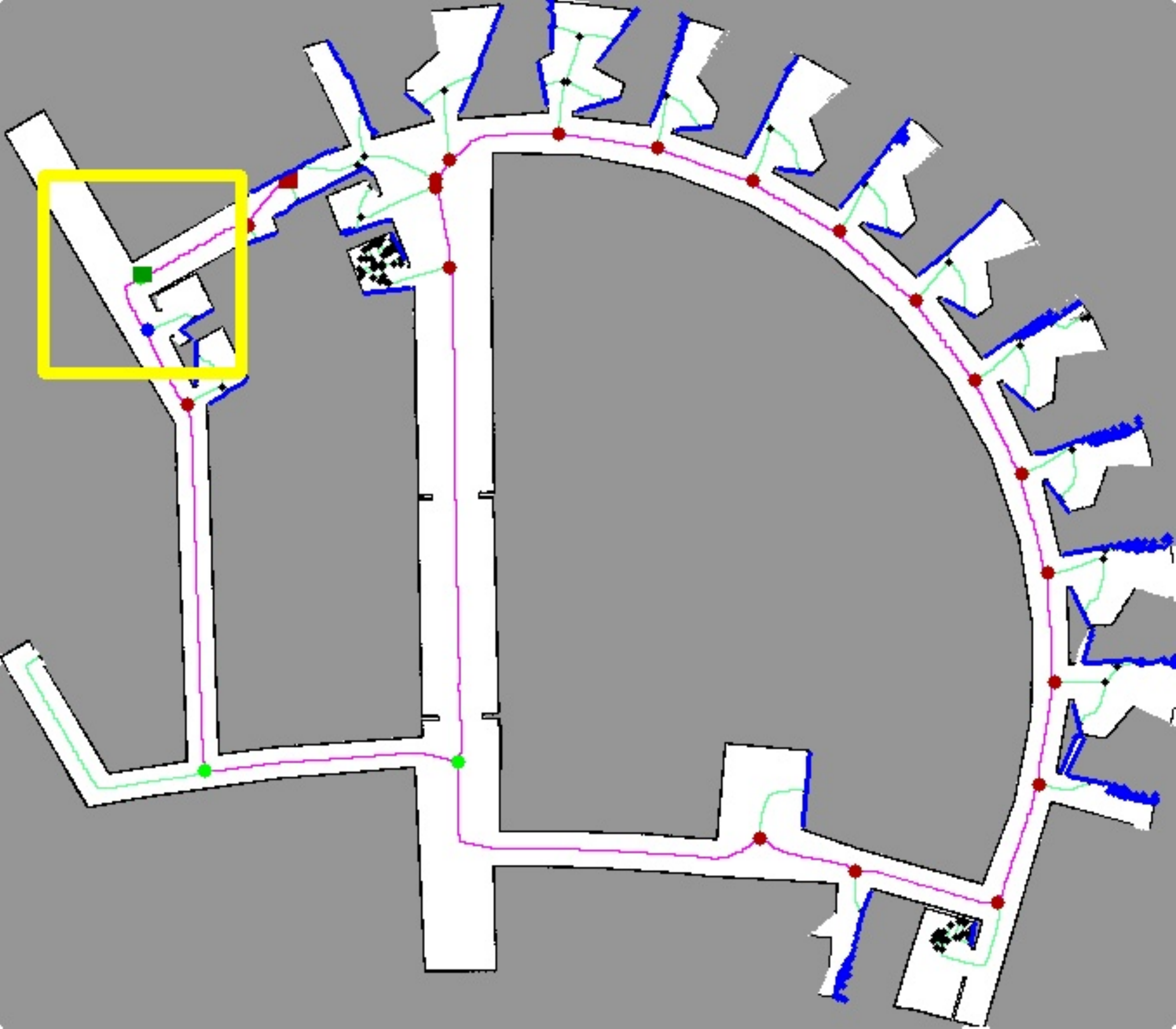}}
\subfigure[Time stamp: 640.7s]{\includegraphics[width=1.65in,height=1.3in]{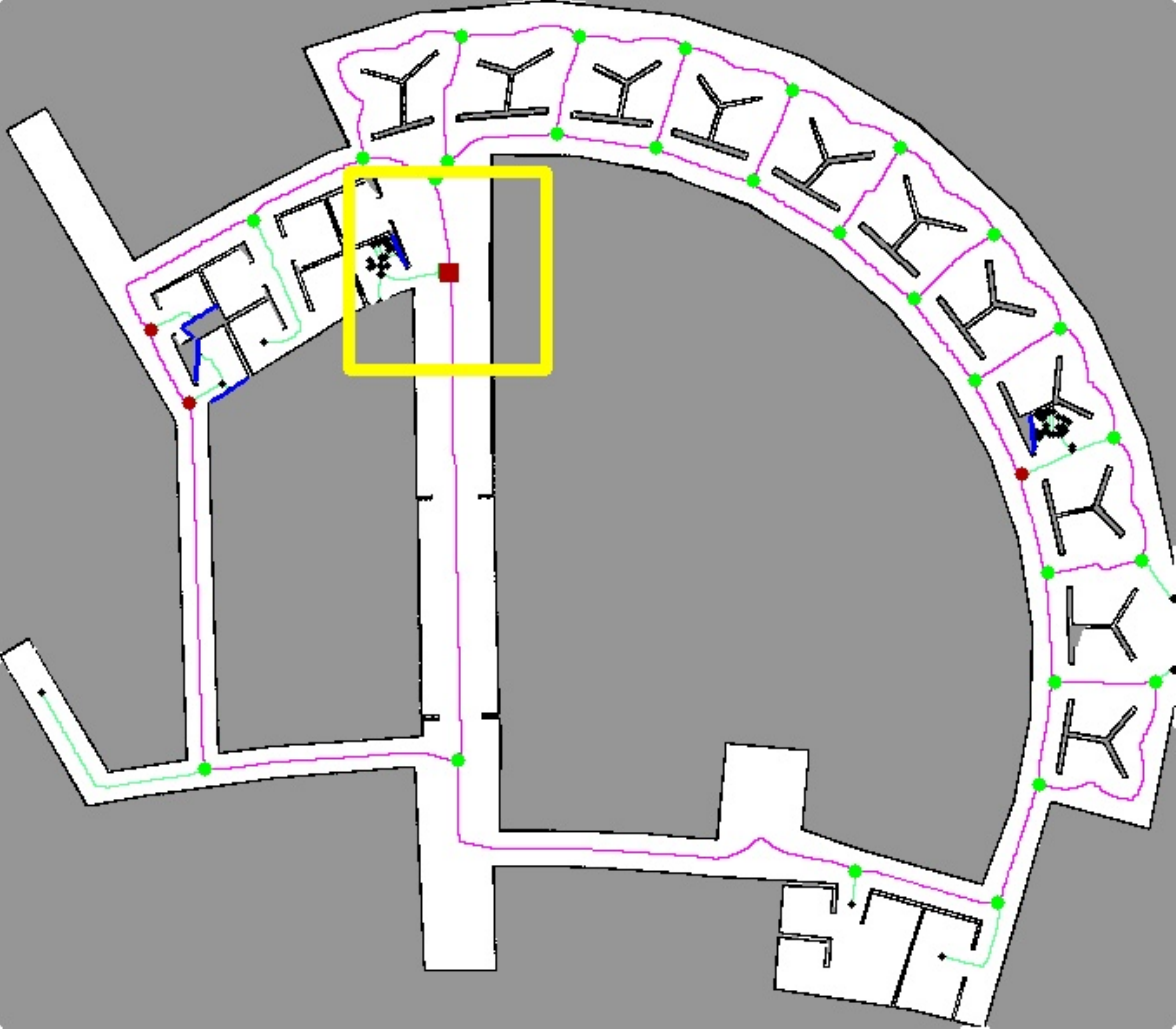}}
\subfigure[Time stamp: 46.7s ]{\includegraphics[width=1.65in,height=1.0in]{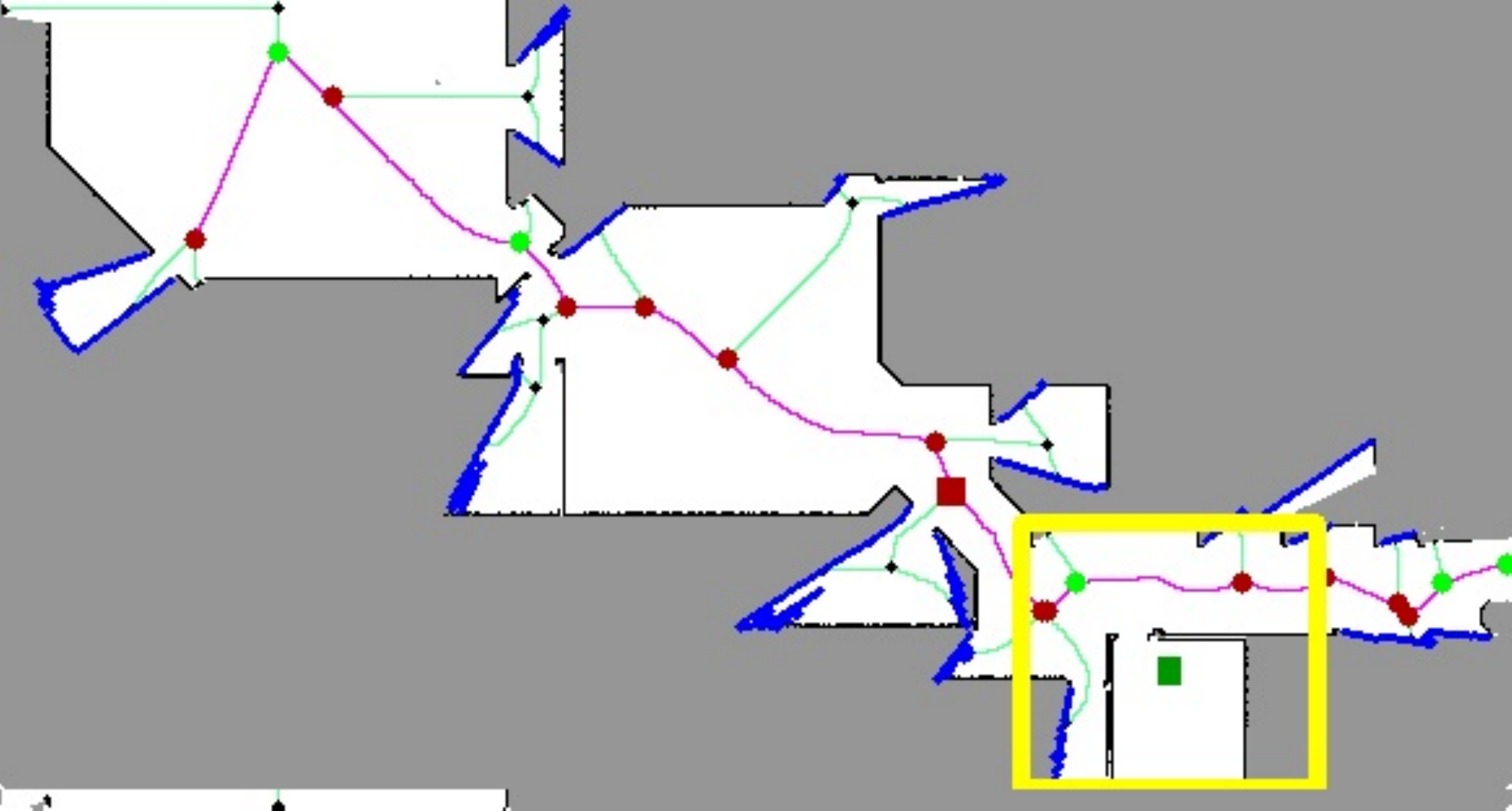}}
\subfigure[Time stamp: 453.1s]{\includegraphics[width=1.65in,height=1.0in]{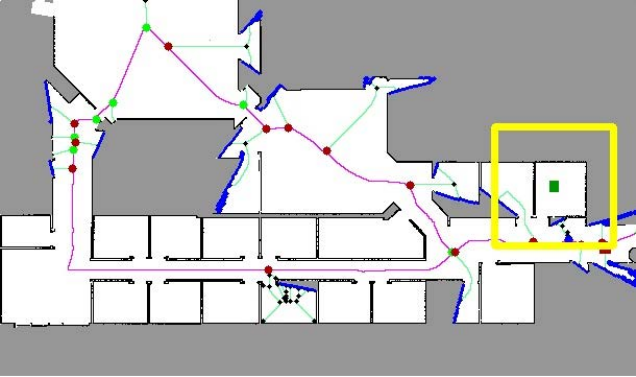}}
\subfigure[Time stamp: 687.3s]{\includegraphics[width=1.65in,height=1.0in]{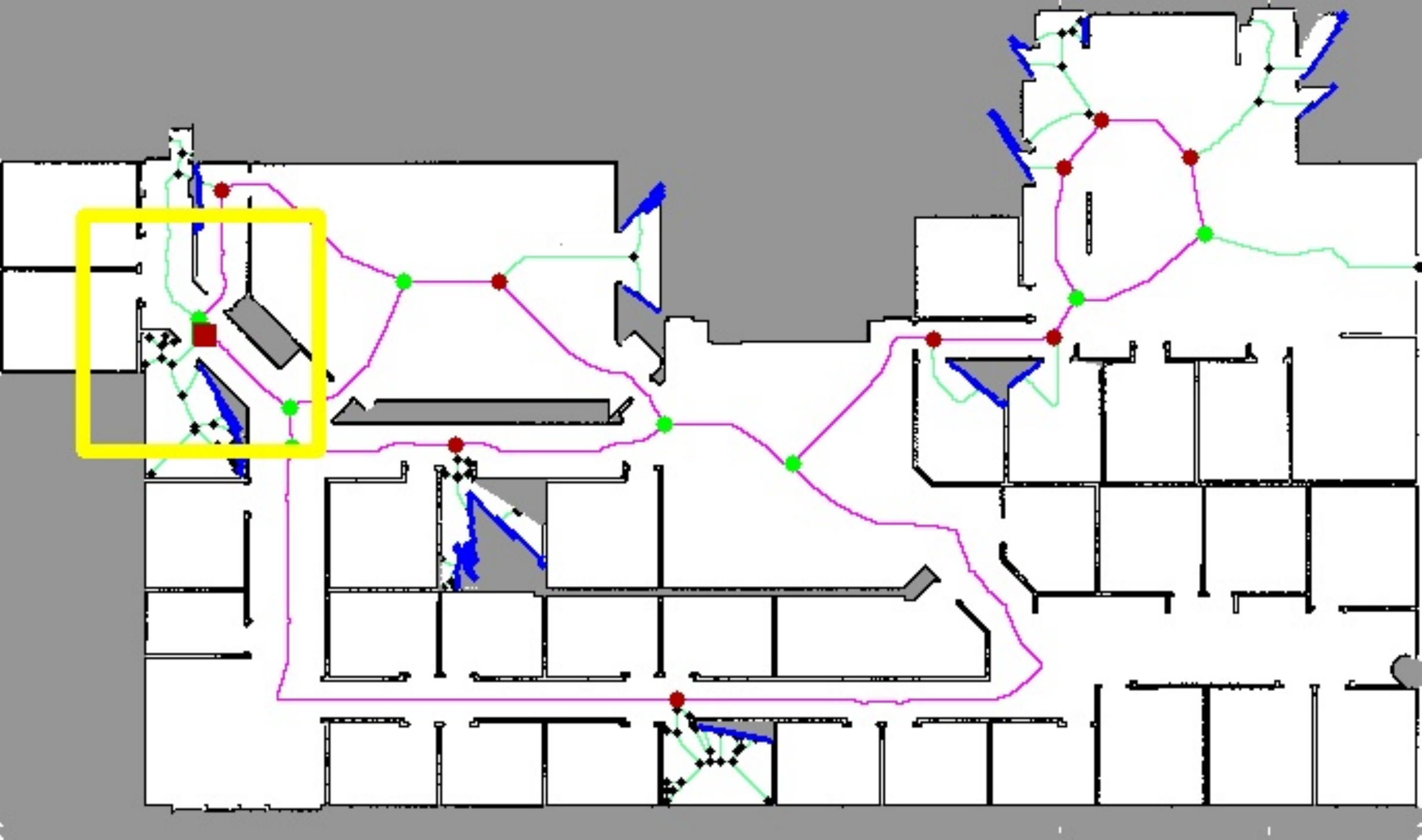}}
\subfigure[Time stamp: 836.8s]{\includegraphics[width=1.65in,height=1.0in]{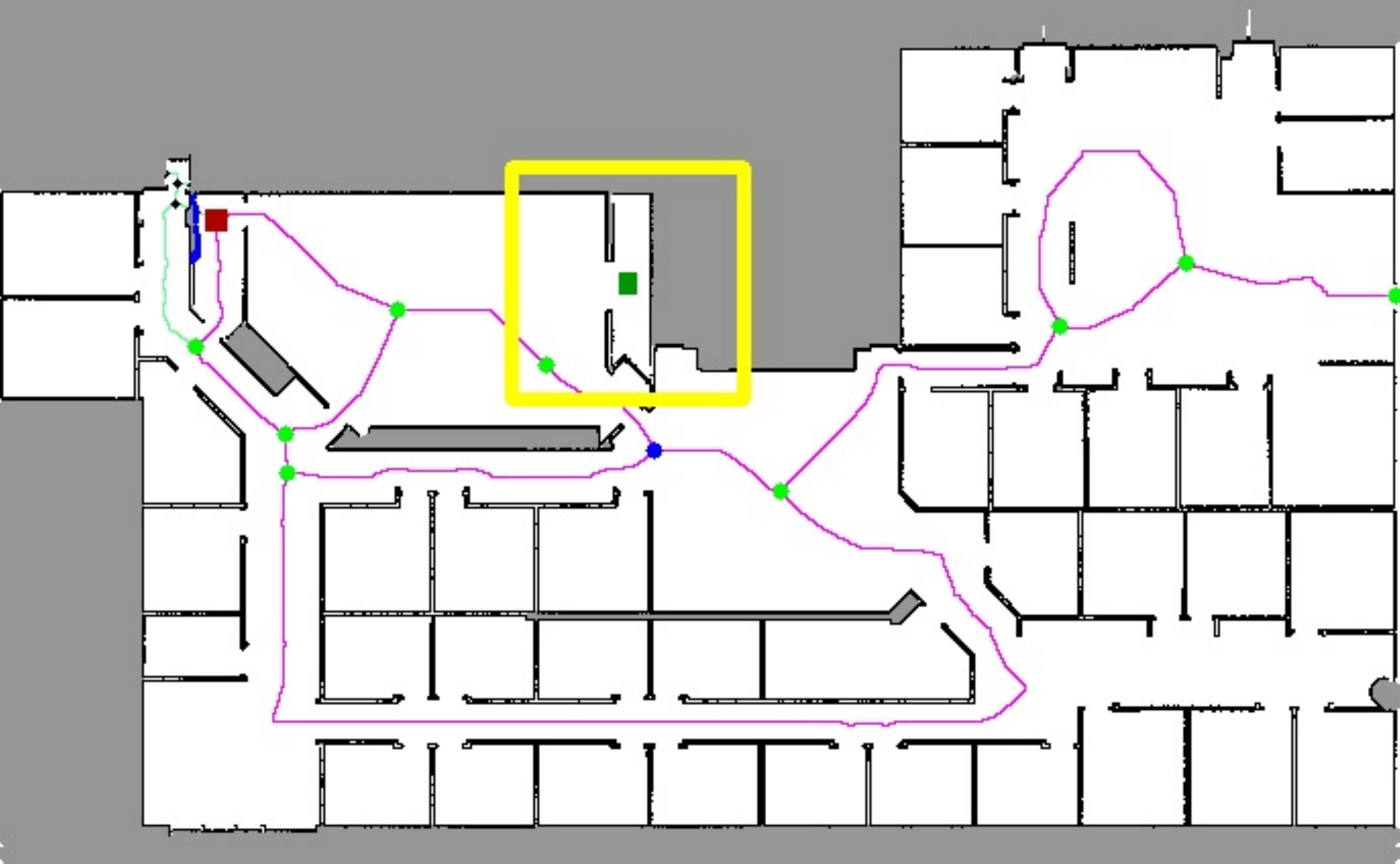}}
}
\caption{Hybrid map representation for simulation test.}
\vspace{-0.4cm}
\label{simulation}
\end{figure*}

The simulation results using the proposed approach are presented in Fig. \ref{simulation}, where (a)-(d) show the exploration process for map model `$Sketched_1$'; and (e)-(h) for map model `$Sketched_2$'. The GVD-based topological nodes and edges are denoted by different coloured dots and lines. As mentioned, the backbone extracted by \textbf{Algorithm \ref{gvd_dfs}} is shown as the red line. The stem nodes are denoted as red and green dots. Red indicates nodes that are associated with frontier information, through the multi-root tree transformation, while green indicates nodes that have no frontier information. At the early stage, the number of red nodes is quite high, but they quickly turn to green nodes as the robot clears frontiers. The current robot pose is represented by a green marker, which is at the center of the yellow box. The yellow box visualizes the local sliding window, while the blue boundaries indicate the list of $\mathcal {F}$, however only the frontier inside the local window will be considered during the `\emph{LocalFrointerDetector}'.

\begin{figure}[tb]
\centering{
\subfigure[Coverage vs time for map $Sketched_1$]{\includegraphics[width=3in]{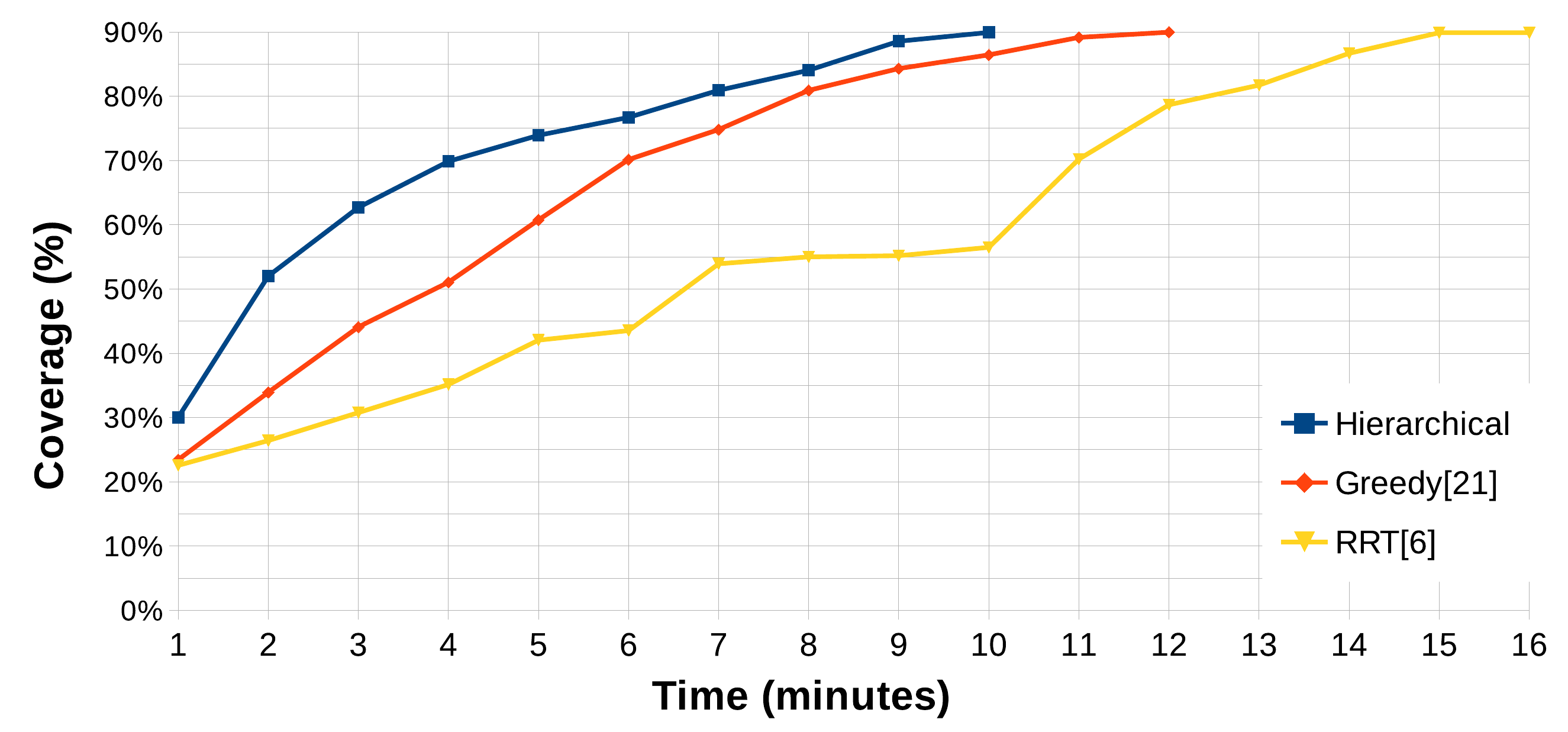}}
\subfigure[Coverage vs time for map $Sketched_2$]{\includegraphics[width=3in]{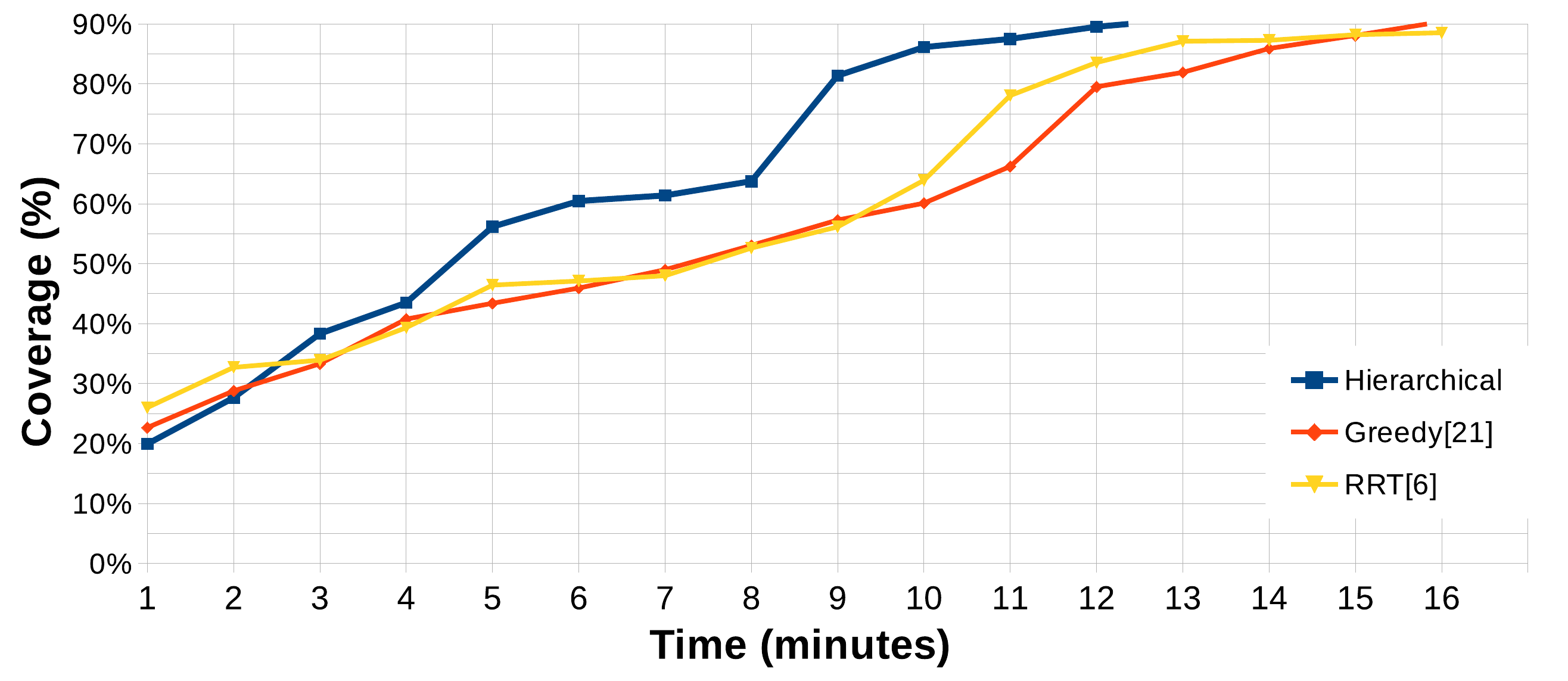}}}
\caption{Comparison of exploration efficiency. The end of each line is the time at which that approach finished, rounded to the nearest minute.}
\vspace{-0.4cm}
\label{timing}
\end{figure}

To further evaluate the advantage of the proposed approach, a graph revealing exploration coverage against time is plotted in Fig. \ref{timing}. The data is calculated by taking the average value of 20 simulation runs. To evaluate the robustness of different methods, the exploration performance under four different initial poses (shown by the red arrows in Fig. \ref{gazebo}) has been tested. Obviously, with equivalent time spending, more unknown space can be detected by using the hierarchical strategy than the other two methods, especially in the early stages of exploration. In other words, the exploration coverage grows the fastest by using the proposed approach. The RRT exploration \cite{Umari2017}, consuming high computational resources, shows a poor performance for the first map model due to low sampling efficiency when using the randomized-tree in a complex environment with multiple tight corridors. In Table \ref{comparison}, the time taken to cover $60\%$, $80\%$ and $90\%$ is shown along with the average processing rate of each approach (default rate is $20 hz$). During the simulation, the last $10\%$ coverage for each approach consisted of small frontiers that the robot had difficulty in navigating to. The greedy frontier approach \cite{online}, always explores the closest and biggest frontier but lacks exploration plan with a global vision. As a result it consumes slightly less computational resources, but results in a much longer exploration time for the large-scale map of $Sketched_2$.

\subsection {Experimental Results}
The real robot platform is built based on a differential-drive mobile platform (Pioneer). A Hokuyo UTM-30-LX laser scanner is mounted about \(30cm\) from the ground, which allows the environment around the robot to be scanned. Different exploration algorithms are implemented on the mentioned platform and tested in our office (9th and 13th floor of Fusionopolis, Singapore). As both maps consist of long and narrow corridors with shielded cubicles/rooms on both sides, they would be considered challenging for self-exploration. The estimated size of each floor is given in Table \ref{comparison} as well.

\begin{figure*}[htb]
	\centering{
    \subfigure[Local planner activated]
    {\includegraphics[width=1.5in,height=1.4in]{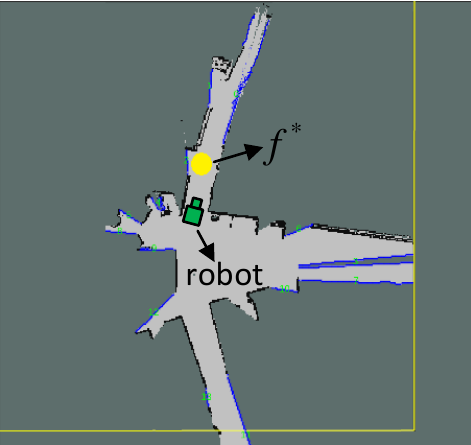}}
    \subfigure[Global planner activated]
    {\includegraphics[width=1.5in,height=1.4in]{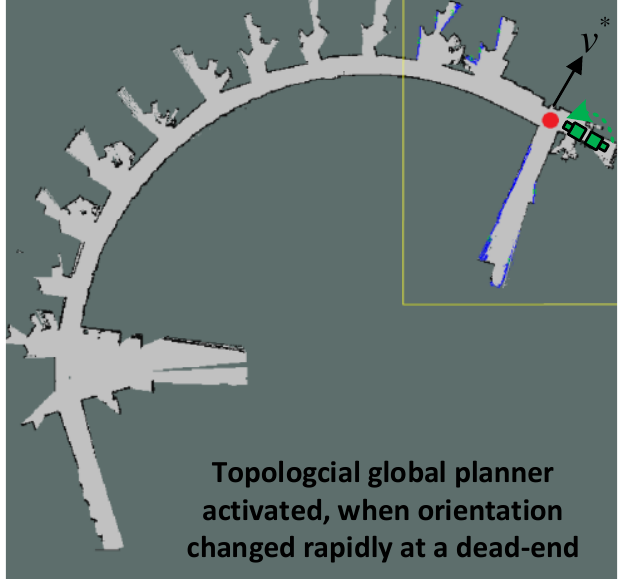}}
    \subfigure[Local planner activated]
    {\includegraphics[width=1.5in,height=1.4in]{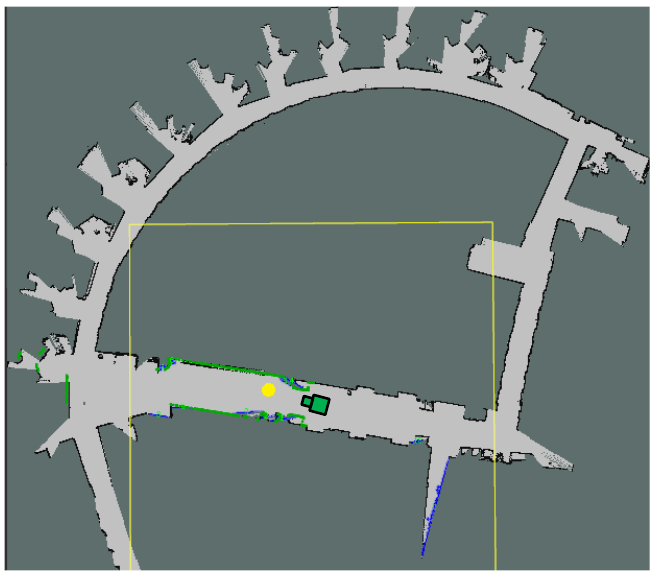}}
    \subfigure[Global planner activated]
    {\includegraphics[width=1.5in,height=1.4in]{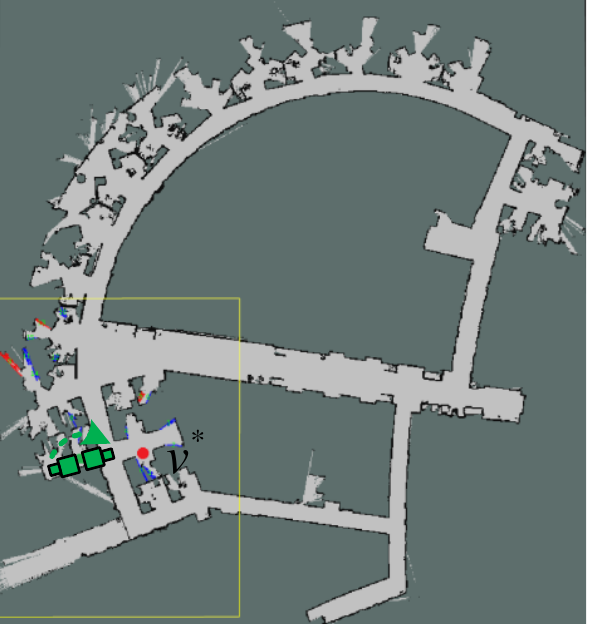}}
    \subfigure[]
    {\includegraphics[width=1.5in,height=1.4in]{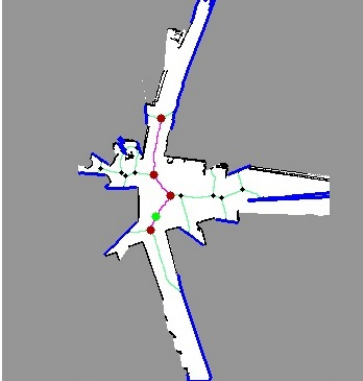}}
    \subfigure[]
    {\includegraphics[width=1.5in,height=1.4in]{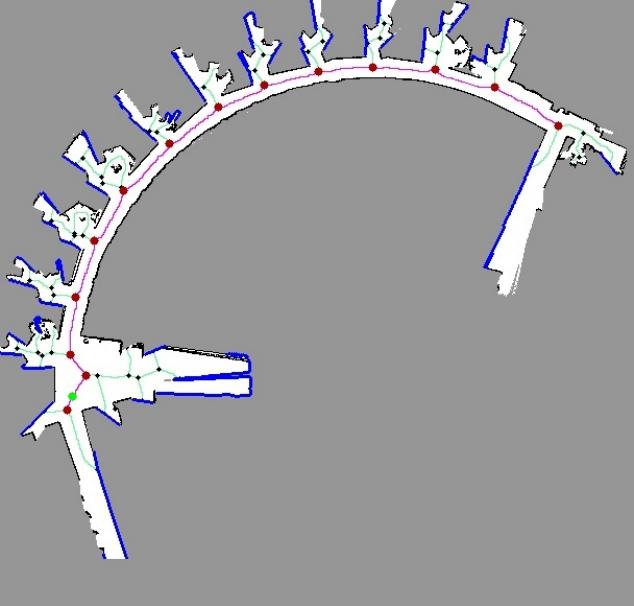}}
    \subfigure[]
    {\includegraphics[width=1.5in,height=1.4in]{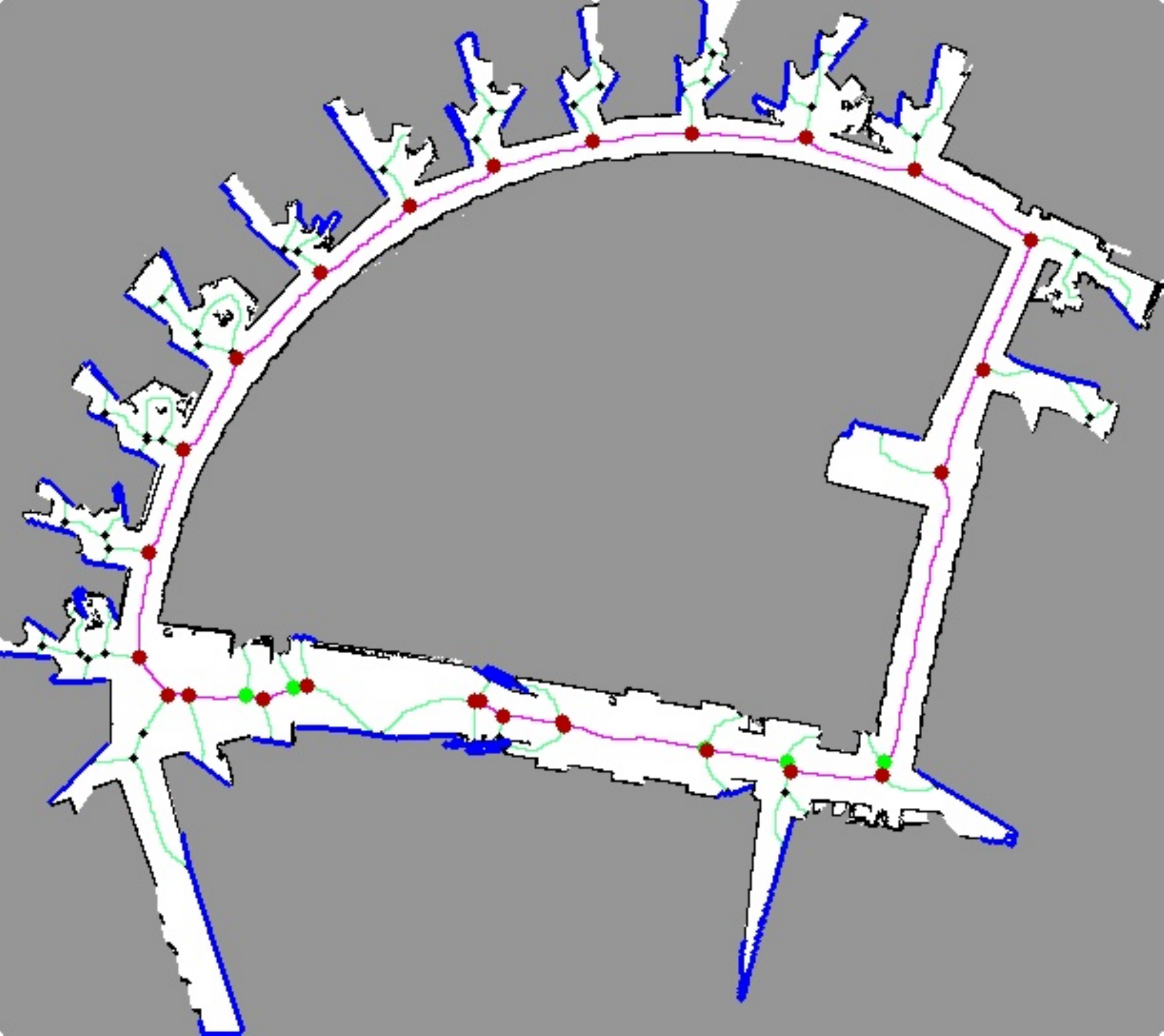}}
    \subfigure[]
    {\includegraphics[width=1.5in,height=1.4in]{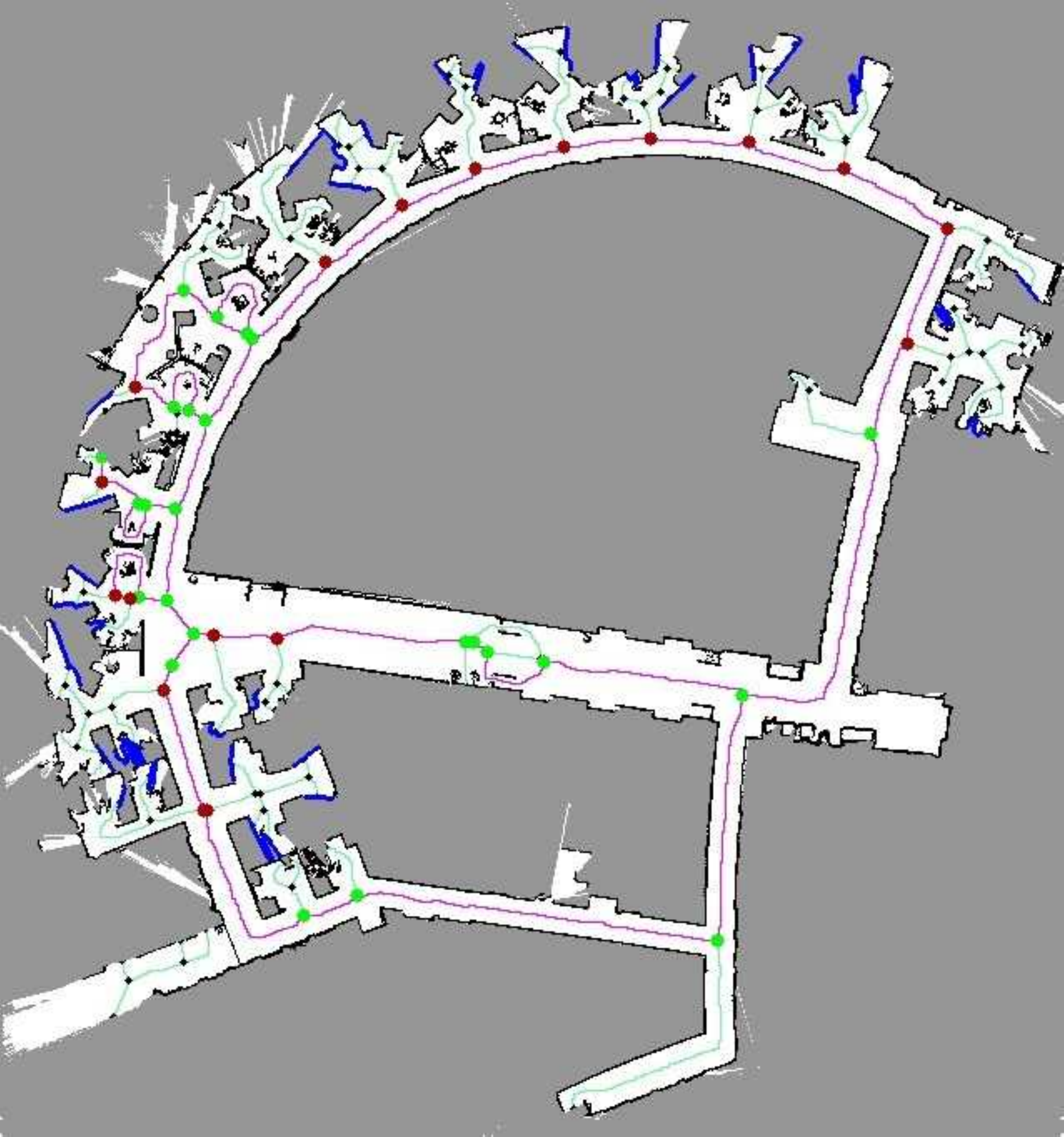}}}
	\caption{Real world experiment by proposed hierarchical exploration in $Office_{9th}$. The exploration process (a)-(d) is demonstrated in `rviz'. The corresponding GVD topological maps are shown as (e)-(h).}
    \vspace{-0.2cm}
	\label{exploration_9th}
\end{figure*}

\begin{figure*}[htb]
	\centering{
    \subfigure[Local planner activated]
    {\includegraphics[width=1.5in,height=1.4in]{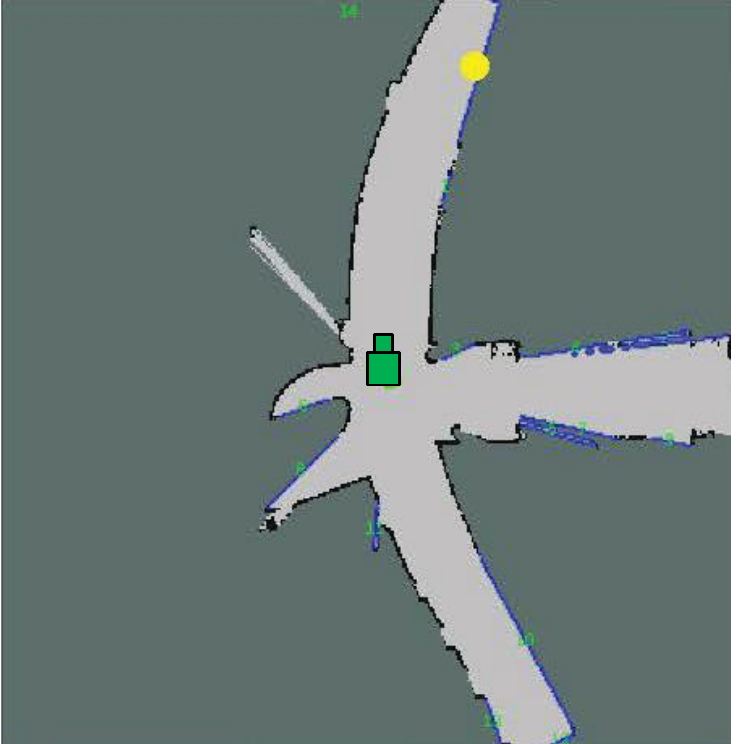}}
    \subfigure[Local planner activated]
    {\includegraphics[width=1.5in,height=1.4in]{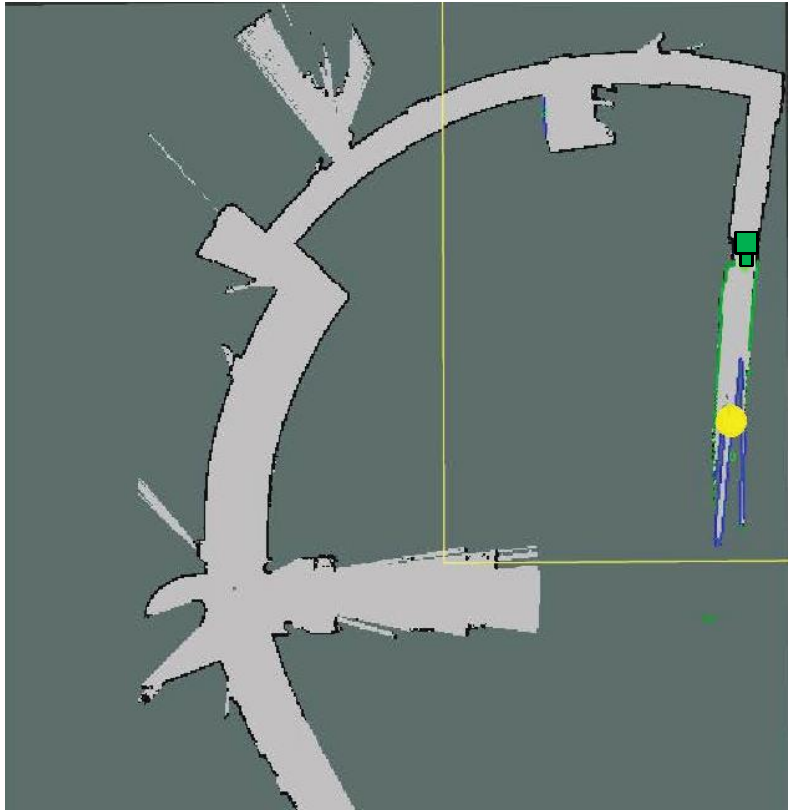}}
    \subfigure[Global planner activated]
    {\includegraphics[width=1.5in,height=1.4in]{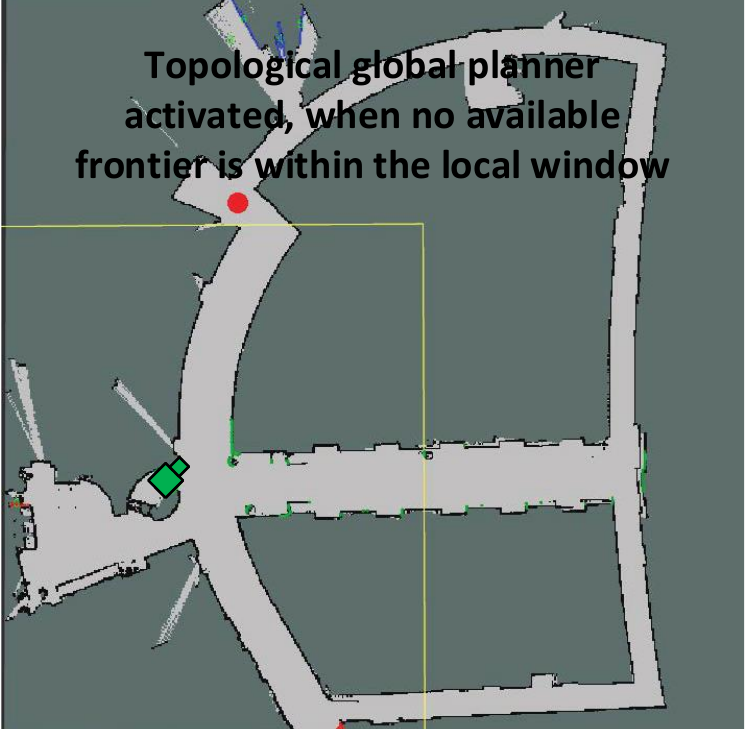}}
    \subfigure[Local planner activated]
    {\includegraphics[width=1.5in,height=1.4in]{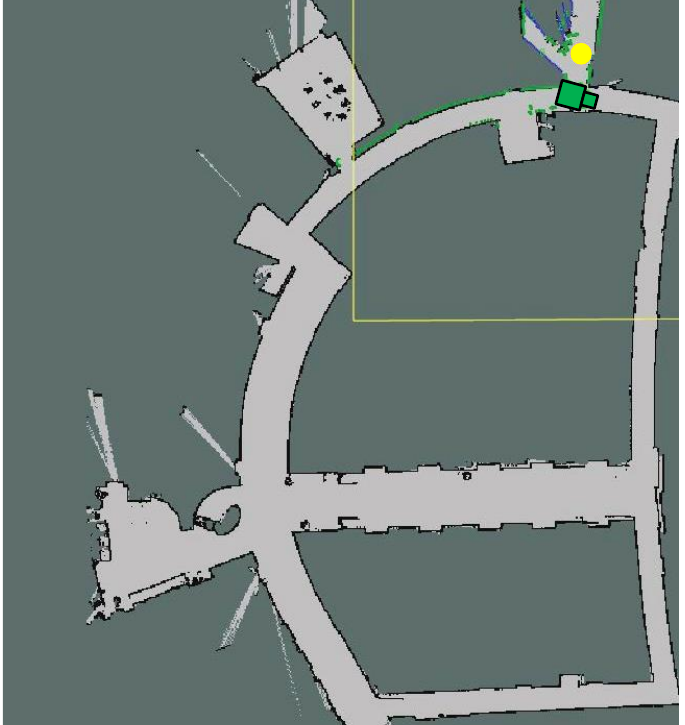}}
    \subfigure[]
    {\includegraphics[width=1.5in,height=1.4in]{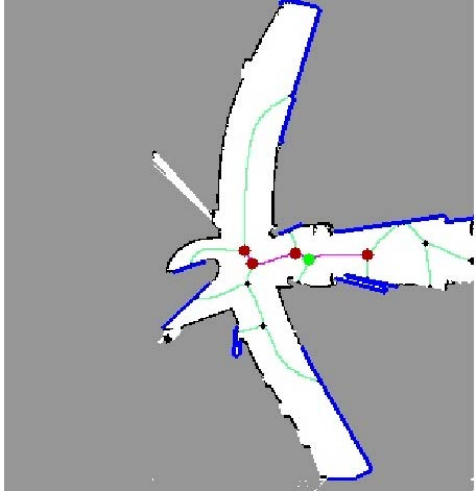}}
    \subfigure[]
    {\includegraphics[width=1.5in,height=1.4in]{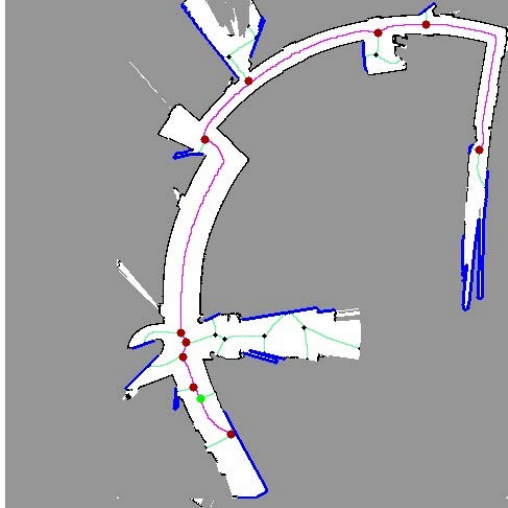}}
    \subfigure[]
    {\includegraphics[width=1.5in,height=1.4in]{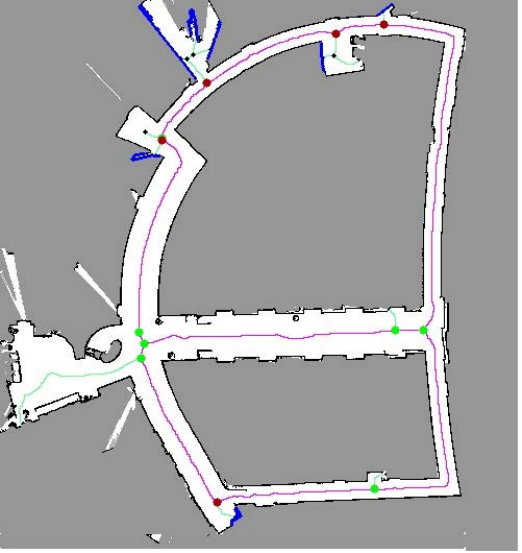}}
    \subfigure[]
    {\includegraphics[width=1.5in,height=1.4in]{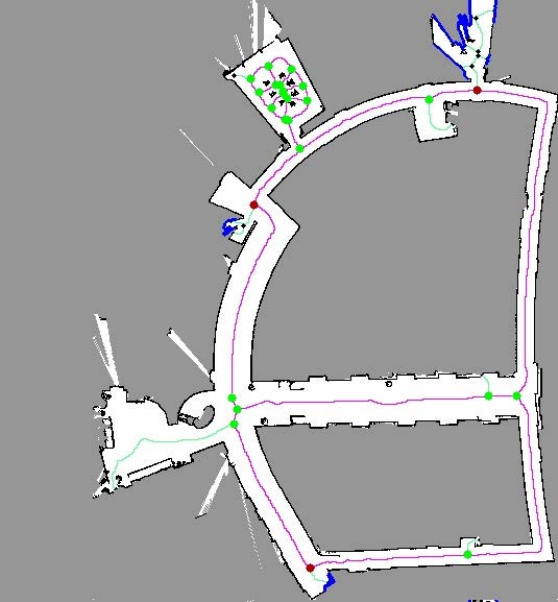}}}
	\caption{Real world experiment by proposed hierarchical exploration in $Office_{13th}$. The exploration process (a)-(d) is demonstrated in `rviz'. The corresponding GVD topological maps are shown as (e)-(h).}
    \vspace{-0.4cm}
	\label{exploration_13th}
\end{figure*}

\begin{center}
\begin{table*}[htb]
\caption{Comparison table for different exploration approaches}
\begin{center}
\setlength{\tabcolsep}{0.2
mm}{
\begin{tabular}{|c|c|c|c|c|c|c|c|}
  \hline
  Map model   & Data type & Size $(m^2)$ & Approach  & $Time\_spend_{60 \%}(sec)$ & $Time\_spend_{80 \%}(sec)$ & $Time\_spend_{90 \%}(sec)$  & $Rate(hz)$  \\ \hline
   \multirow{3}{*}{$Sketched_1$}       &  \multirow{3}{*}{Simulated} & \multirow{3}{*}{642}
   & {Hierarchical}             & {\textbf{168}}   & {\textbf{358}}   & {\textbf{601}}   & {18.5}  \\ \cline{4-8}
   & & & {Greedy}\cite{online}& {297}            & {482}     & {658}                & {{19.8}}  \\ \cline{4-8}
   & & & {RRT}\cite{Umari2017}  & {626}            & {737}     & {924}                & {13.5}  \\ \hline

   \multirow{3}{*}{$Sketched_2$}    &  \multirow{3}{*}{Simulated} & \multirow{3}{*}{997}
   & {Hierarchical}             & {\textbf{384}}   & {\textbf{534}}   & {\textbf{753}}   & {19.1}  \\ \cline{4-8}
   & & & {Greedy}\cite{online}& {566}            & {721}            & {953}            & {{19.6}}  \\ \cline{4-8}
   & & & {RRT}\cite{Umari2017}  & {575}            & {699}            & {981}            & {13.8}  \\ \hline

   \multirow{3}{*}{$Office_{9th}$}  &  \multirow{3}{*}{Real} & \multirow{3}{*}{536}
   & {Hierarchical}             & {\textbf{213}}   & {\textbf{778}}    & {\textbf{906}}   & {16.6}  \\ \cline{4-8}
   & & & {Greedy}\cite{online}& {376}            & {1046}     & {Failed}                & {{17.3}}  \\ \cline{4-8}
   & & & {RRT}\cite{Umari2017}  & {478}            & {Failed}     & {Failed}              & {10.9}  \\ \hline

   \multirow{3}{*}{$Office_{13th}$} &  \multirow{3}{*}{Real} & \multirow{3}{*}{321}
   & {Hierarchical}             & {\textbf{127}}   & {\textbf{208}}    & {276}          & {14.9}  \\ \cline{4-8}
   & & & {Greedy}\cite{online}& {181}            & {246}             & {{273}}        & {{16.4}}  \\ \cline{4-8}
   & & & {RRT}\cite{Umari2017}  & {314}            & {443}             & {Failed}       & {11.4}  \\ \hline

\end{tabular}}
\end{center}
\vspace{-0.3cm}
\label{comparison}
\end{table*}
\end{center}

The experimental results using the proposed approach are illustrated in Fig. \ref{exploration_9th} and Fig. \ref{exploration_13th} respectively. The exploration process, including map building and topological graph construction, is gradually demonstrated from the starting position in (a) until most of the space has been explored in (d). Subfigures (a)-(d) are snapshots of the visualization tool `rviz' during self-exploration. The corresponding topological graph $G$ shown in subfigures (e)-(h) updates with respect to the robot position, which is indicated by the green marker. Although graph $G$ is shown in each step, the `\emph{GVDTopologicalPlanner}' may not be activated each time. Frontiers $\mathcal {F}$ within the local window (yellow box) denoted by blue boundaries are used to determine $f^*$. By running the `\emph{LocalFrontierDetector}' the navigational coordinates for the desired frontier is selected and is displayed as a yellow dot. Whenever `\emph{GVDTopologicalPlanner}' is activated, the exploration goal $v^*$ can be determined via the four steps mentioned in Section \ref{methodology}.C. As a result, the global planner was activated and the exploration goal was marked by a red dot in both figures when the robot reached an dead-end in Fig. \ref{exploration_9th}(b) or when no available frontier existed in the local window in Fig. \ref{exploration_13th}(c).

\begin{figure}[tb]
\centering{
\subfigure[Greedy frontier exploration in $Office_{9th}$, time\_stamp=380 sec]{\includegraphics[width=1.6in,height=1.4in]{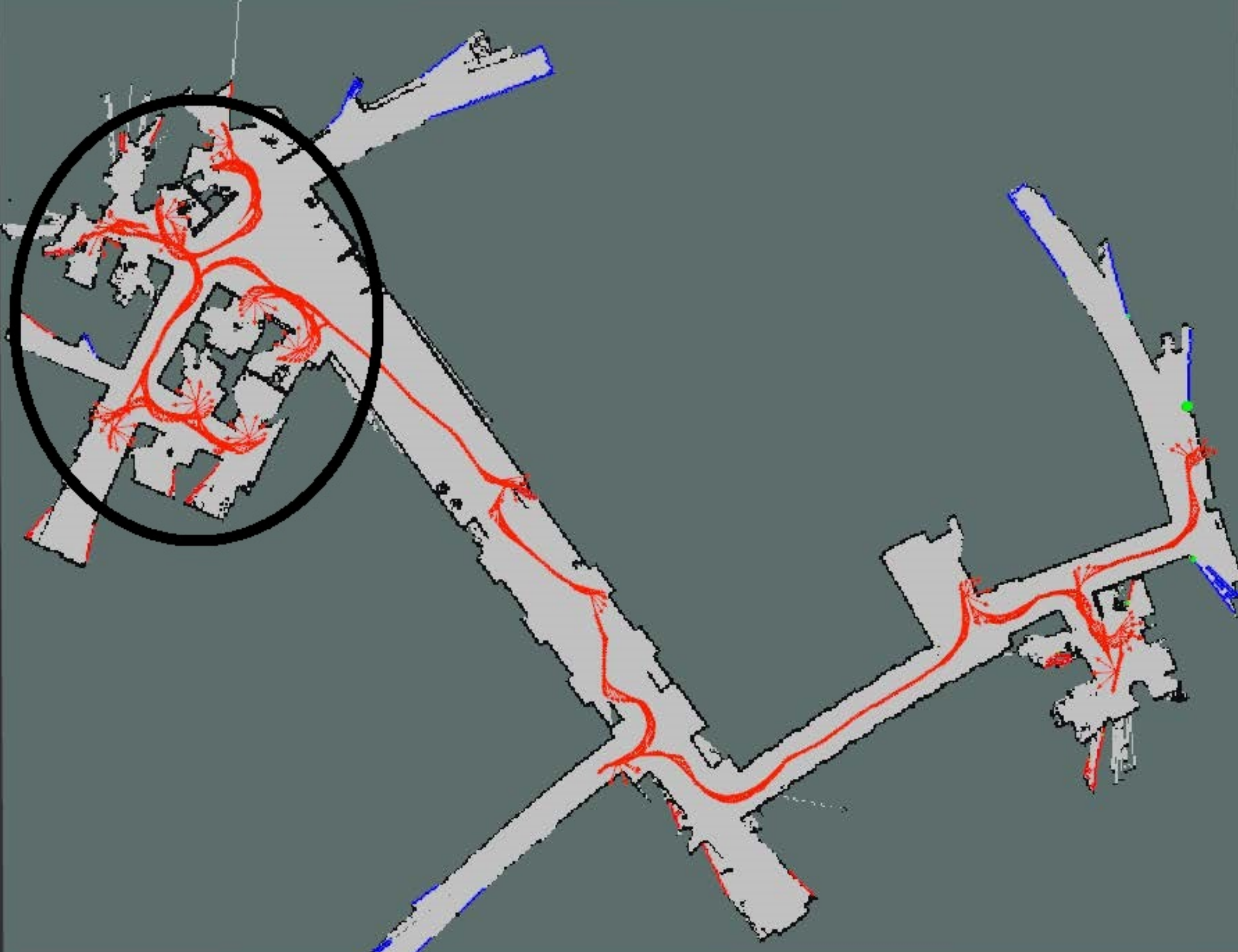}}
\subfigure[Hierarchial exploration in $Office_{9th}$, time\_stamp=380 sec]{\includegraphics[width=1.5in,height=1.4in]{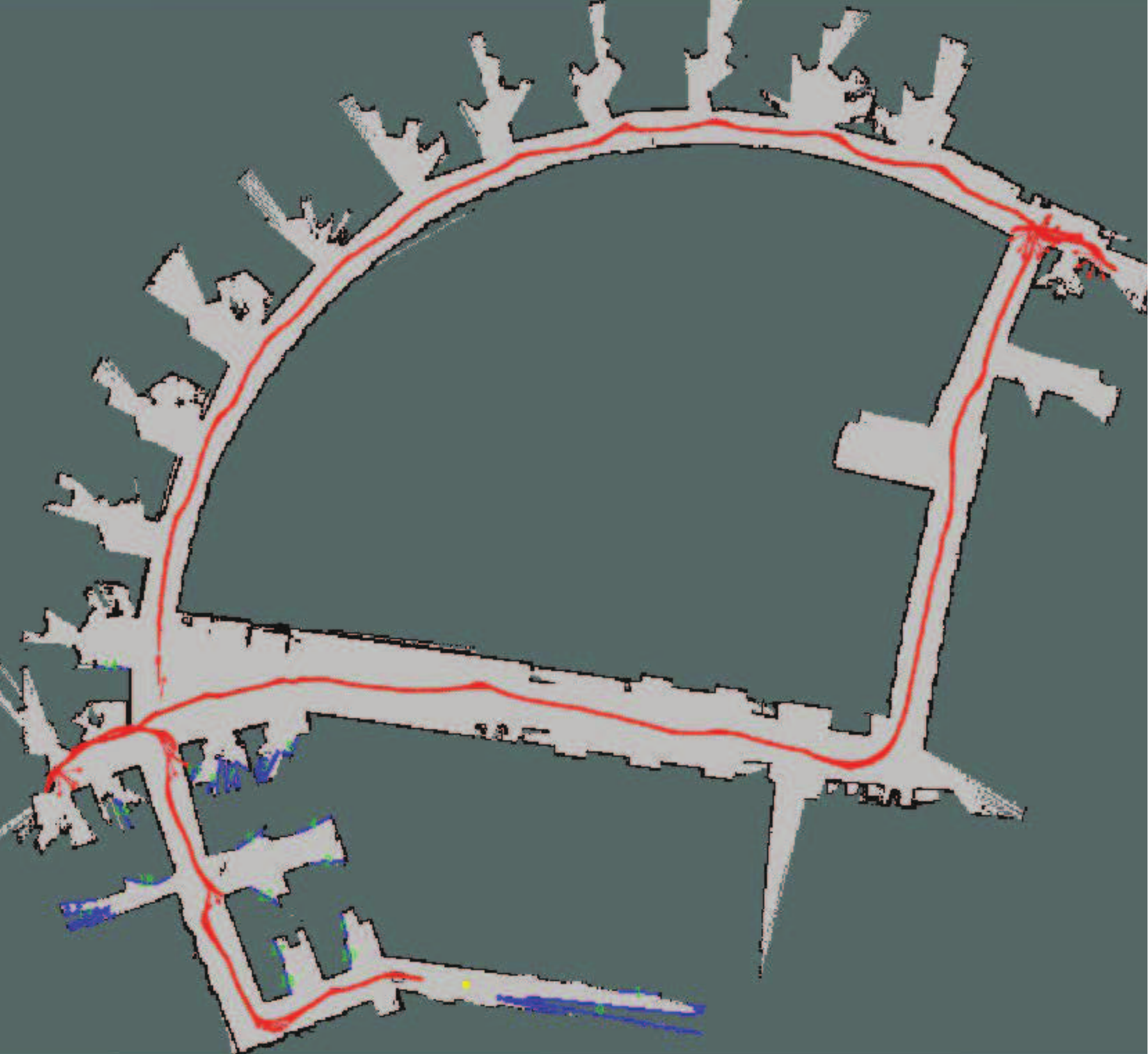}}}
\caption{An example of exploration trajectory using the hierarchial strategy vs. greedy frontier method \cite{online} under the same initial conditions}
\vspace{-0.4cm}
\label{standard_perform}
\end{figure}

The performance of real robot exploration is summarized in Table \ref{comparison}. In both cases, the time used to cover $60\%$ and $80\%$ of the whole map is significantly less when using the hierarchial approach. For the greedy approach, a similar exploration speed can be achieved to cover $90\%$ of the map in a much simpler environment of $Office_{13th}$, where the space is relatively open and regular. However, the same approach failed to explore the more complex environment $Office_{9th}$ up to $90\%$ as it was travelling back and forth repeatedly over explored location. A example of exploration trajectory performed by the greedy frontier is shown in Fig. \ref{standard_perform}(a), where quite a few back and forth movements can be observed inside the highlighted region, while no such behaviour can be observed under the proposed framework in Fig. \ref{standard_perform}(b). Note that the extra CPU usage by taking the hierarchial architecture is within an acceptable range. Therefore, in terms of time expenditure and map coverage, the experimental results validate the superior exploration efficiency of the hierarchical exploration strategy compared to the greedy method in complex office environments. In fact, an additional advantage of using the hierarchial approach can be inferred that the stem-preferred strategy could also increase the probability of loop-closure during the SLAM process. Meanwhile, according to Table \ref{comparison}, the RRT exploration method \cite{Umari2017} showing a poor search coverage and low processing rate performed less inefficiently than the proposed one as well.

\section{Conclusion and Discussion}
\label{conclusion}

In this paper, by combining the benefits of metric and topological map, an exploration strategy using hybrid map representation is proposed to solve a challenging problem of self-exploration and mapping in a complex environment. Two planning stages are designed to collaborate with each other hierarchically. The lower-level stage prioritizing the local frontier along with the robot motion direction will force the robot to follow the main road (if it is available). The upper-level planner, leveraging on the technique of GVD, is able to make global decisions based on a modified tree data structure called a multi-root tree. By consolidating the frontier information from all leaf nodes, a systematic way of determining the optimal exploration goal can be achieved. The exploration goal determined by either planning stage is assigned to the robot base to work together with the process of SLAM. The proposed approach is evaluated in both simulation and experimental environments by comparing against two other methods. According to the results, the proposed approach achieves the greatest exploration efficiency when exploring in a typical office area. In the future, multi-robot exploration will be considered, such that one robot focuses exploring on the stem road, whilst others deal with the branches simultaneously. In addition, more experiments should be carried out in different types of environments.

\end{document}